%% file: main.tex
\documentclass{article}
\input{preamble}

\input{cref_setup}

\input{macros}

\usepackage[accepted]{icml2025}

\icmltitlerunning{PhantomWiki: On-Demand Datasets for Reasoning and Retrieval Evaluation}

\begin{document}

\twocolumn[
\icmltitle{PhantomWiki: On-Demand Datasets for Reasoning and Retrieval Evaluation}

\icmlsetsymbol{equal}{*}

\begin{icmlauthorlist}
\icmlauthor{Albert Gong}{equal,cor}
\icmlauthor{Kamilė Stankevičiūtė}{equal,cor,cam}
\icmlauthor{Chao Wan}{equal,cor}
\icmlauthor{Anmol Kabra}{equal,cor}

\icmlauthor{Raphael Thesmar}{cor}
\icmlauthor{Johann Lee}{cor}
\icmlauthor{Julius Klenke}{cor}
\icmlauthor{Carla P. Gomes}{cor}
\icmlauthor{Kilian Q. Weinberger}{cor}

\end{icmlauthorlist}

\icmlaffiliation{cor}{Department of Computer Science, Cornell University, Ithaca, New York, USA}
\icmlaffiliation{cam}{Department of Computer Science and Technology, University of Cambridge, Cambridge, UK}

\icmlcorrespondingauthor{Albert Gong}{agong@cs.cornell.edu}
\icmlcorrespondingauthor{Kamilė Stankevičiūtė}{ks830@cam.ac.uk}
\icmlcorrespondingauthor{Chao Wan}{cw862@cornell.edu}
\icmlcorrespondingauthor{Anmol Kabra}{anmol@cs.cornell.edu}

\icmlkeywords{Machine Learning, ICML}

\vskip 0.3in
]

\printAffiliationsAndNotice{\icmlEqualContribution} %

\input{sections/abstract}

\input{sections/intro}
\input{sections/related}
\input{sections/dataset}
\input{sections/main_results}
\input{sections/reasoning}
\input{sections/retrieval}
\input{sections/finetuning_on_pw}
\input{sections/future}

\section*{Software and Data}
The source code for this work can be found at \href{https://github.com/kilian-group/phantom-wiki}{github.com/kilian-group/phantom-wiki} and via \texttt{pip install phantom-wiki}, and the sample HuggingFace datasets are available at \href{https://huggingface.co/datasets/kilian-group/phantom-wiki-v1}{kilian-group/phantom-wiki-v1}.

\ifdefined\isaccepted
\section*{Acknowledgements}

KS gratefully acknowledges support from AstraZeneca. 
CW is supported by the National Science Foundation (NSF) OAC-2118310 and NSF-1934714 grant. 
This work was partially supported by funding from NewYork-Presbyterian for the NYP-Cornell Cardiovascular AI Collaboration, the National Institute of Food and Agriculture (USDA/NIFA), the Air Force Office of Scientific Research (AFOSR), and a Schmidt AI2050 Senior Fellowship, a Schmidt Sciences program.
The authors thank anonymous reviewers for their helpful feedback on this work.
\fi

\section*{Impact Statement}

By leveraging context-free grammars and Prolog, \ours is able to generate large, durable and challenging datasets without using LLMs.
The datasets have low computational, monetary, and environmental cost and our open-source framework is accessible to any user.

Since \ours randomly generates datasets that do not reference any existing data, the evaluation benchmark is resistant to data leakage and memorization while training.
The approach of publishing a dataset generation \textit{procedure} rather than a fixed dataset also encourages better research practices (by using fresh datasets instead of overfitting to a single instance), and enables a more accurate evaluation of model performance. 
Since we do not use any personal data, use of \ours does not have any privacy concerns.

\bibliography{refs}
\bibliographystyle{icml2025}

\input{appendix}

\end{document}

%% file: preamble.tex
\usepackage{microtype}
\usepackage{graphicx}
\usepackage{array}
\usepackage{subfigure}
\usepackage{booktabs} %
\usepackage{enumitem} %
\usepackage{listings}
\usepackage{quoting}

\usepackage{hyperref}
\usepackage[dvipsnames]{xcolor}
\hypersetup{
	colorlinks,
	linkcolor={red!50!black},
	citecolor={blue!50!black},
	urlcolor={blue!80!black}
}
\usepackage{xurl}

\usepackage{amsmath}
\usepackage{amssymb}
\usepackage{mathtools}
\usepackage{amsthm}

\usepackage{ksmath}

\usepackage[colorinlistoftodos,prependcaption,textsize=tiny]{todonotes}
\providecommand{\tdotoggle}{1}
 \ifnum\tdotoggle=1
 \fi

\usepackage{xspace}
\usepackage{multirow}

\usepackage{colortbl}
\usepackage{subcaption}
\usepackage{nicematrix}
\usepackage{tcolorbox}
\usepackage{float}

%% file: cref_setup.tex
\Crefformat{footnote}{#2\footnotemark[#1]#3}

%% file: macros.tex
\newcommand{\inbraces}[1]{\left\{#1\right\}}

\newcommand{\set}[1]{\inbraces{#1}}

\def\orange#1{\textcolor{orange}{{#1}}}

\def\red#1{\textcolor{red}{{#1}}}

\definecolor{Gred}{RGB}{219, 50, 54}
\definecolor{Ggreen}{RGB}{60, 186, 84}
\definecolor{Gblue}{RGB}{72, 133, 237}
\definecolor{Gyellow}{RGB}{247, 178, 16}
\definecolor{ToCgreen}{RGB}{0, 128, 0}
\definecolor{myGold}{RGB}{231,141,20}
\definecolor{myBlue}{rgb}{0.19,0.41,.65}
\definecolor{myPurple}{RGB}{175,0,124}

\newcommand{\mytodo}[1]{\ifnum\tdotoggle=1{#1}\fi}

\newcommand{\tableoftodos}{\ifnum\tdotoggle=1 \listoftodos[Comments/To Do's] \fi}

\newcommand{\ours}{\text{PhantomWiki}\xspace}

\newcommand{\zeroshot}{\textsc{ZeroShot}\xspace}
\newcommand{\CoT}{\textsc{CoT}\xspace}
\newcommand{\simpleprompting}{\text{in-context prompting}\xspace}
\newcommand{\ragprompting}{\text{RAG prompting}\xspace}
\newcommand{\agenticprompting}{\text{agentic prompting}\xspace}
\newcommand{\Simpleprompting}{\text{In-Context prompting}\xspace}
\newcommand{\Agenticprompting}{\text{Agentic prompting}\xspace}

\newcommand{\zeroshotsimple}{\hyperref[sub:zeroshot_simple]{\color{black} {\textsc{ZeroShot}}}\xspace}
\newcommand{\zeroshotrag}{\hyperref[sub:zeroshot_rag]{\color{black} {\textsc{ZeroShot-RAG}}}\xspace}
\newcommand{\cotsimple}{\hyperref[sub:cot_simple]{\color{black} {\textsc{CoT}}}\xspace}
\newcommand{\cotrag}{\hyperref[sub:cot_rag]{\color{black} {\textsc{CoT-RAG}}}\xspace}
\newcommand{\react}{\hyperref[sub:react]{\color{black}{\textsc{ReAct}}}\xspace}

\newcommand{\retriever}{\textsc{UAE-Large-V1}\xspace}

\newcommand{\gpt}{GPT-4o\xspace}
\newcommand{\gemini}{Gemini-1.5-Flash\xspace}
\newcommand{\llama}{Llama-3.3-70B\xspace}
\newcommand{\deepseek}{DeepSeek-R1-32B\xspace}

\newcommand{\braces}[1]{\left\{ #1 \right \}}

\def\tbf#1{\textbf{#1}}

%% file: sections/abstract.tex
\begin{abstract}
High-quality benchmarks are essential for evaluating reasoning and retrieval capabilities of large language models (LLMs).
However, curating datasets for this purpose is not a permanent solution as they are prone to data leakage and inflated performance results. 
To address these challenges, we propose \ours: a pipeline to generate unique and factually consistent document corpora with diverse question-answer pairs. 
Unlike prior work, \ours is neither a fixed dataset, nor is it based on any existing data. 
Instead, a new \ours instance is generated on demand for each evaluation. 
We vary the question difficulty and corpus size to disentangle reasoning and retrieval capabilities respectively, and find that \ours datasets are surprisingly challenging for frontier LLMs. Thus, we contribute a scalable and data leakage-resistant framework for disentangled evaluation of reasoning, retrieval, and tool-use abilities.
\end{abstract}

%% file: sections/intro.tex
\section{Introduction}
\label{sec:intro}

Designing agents that can perform complex reasoning while interfacing with a large-scale, dynamic corpus---like Wikipedia---is a long-standing goal in the field of natural language processing \citep{feldman2019multi,min2019discrete}.
Such a goal may be within reach given the impressive capabilities of recent language models, which are all trained on internet-scale data.
For example, the ability of LLMs to solve math problems on GSM8K \citep{cobbe2021training} and mathematical olympiads \cite{alphaproof2024} could bode well for agents to answer highly quantitative questions. 
On benchmarks like DROP \citep{dua2019drop} and MMLU \citep{hendrycks2020measuring}, LLMs demonstrate advanced reading comprehension and general reasoning capabilities, both necessary for intelligent agents.
When augmented with retrievers \citep{muennighoff2022mteb} and tools \citep{patil2023gorilla}, language models seem to already possess a strong ability for accessing external datastores and knowledge bases.

However, it is unclear to what extent these models rely on their internal knowledge, which can easily become outdated, versus their reasoning and retrieval abilities. 
For example, consider the question ``\emph{What is the date of birth of Wolfgang Amadeus Mozart?}''.
Since this fact is contained within LLM pre-training data, asking LLMs this question cannot provide reliable insight on whether the answer was deduced, retrieved or recalled. 
At the same time, existing approaches that perturb Wikipedia facts \citep{cohen2024evaluating,meng2022locating,elazar2021measuring, ho2020constructing} to construct new question-answer pairs face challenges of ensuring factual consistency across articles.
For example, changing Mozart's date of birth to 2025 would also require modifying Beethoven's article to erase the fact that Beethoven might have met Mozart in 1787!

One could hope to isolate reasoning from factual knowledge using mathematical or logical reasoning benchmarks. Unfortunately, such benchmarks are not entirely reliable as indicators of reasoning performance either.
On GSM8K, a dataset of grade school math problems, \citet{mirzadeh2024gsm} report that frontier models perform significantly worse with minor or even meaningless alterations to the test data---indicating these models are vulnerable to overfitting at best and exact memorization at worst. 
To ensure fair comparison, LLMs need to be evaluated in a way that does not depend on any particular dataset instance.

\begin{figure}[t!]
    \centering
    \includegraphics[width=\linewidth]{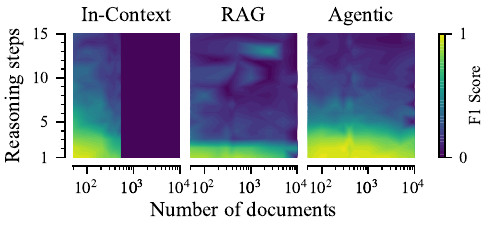}
    \caption{%
    \tbf{Evaluating LLM capabilities with \ours.} We tune the reasoning and retrieval difficulty by the number of reasoning steps and documents, respectively.
    In this representative example of \llama (other LLMs being similar), the top-right regions in all settings grow darker, indicating that the F1 score plummets as both reasoning and retrieval complexity increase.
    Note that in-context prompting has a sharp cut-off along the $x$-axis at \llama's context length limit.
    }
    \label{fig:size-difficulty-f1} 
\end{figure}

Following this philosophy, we develop \emph{PhantomWiki}. 
At the click of a button, \ours generates a synthetic, fictional universe of characters along with a set of facts about them. 
We reflect these facts in a large-scale corpus of templated articles, mimicking the style of fan-wiki websites. 
Then we generate question-answer pairs about the universe, encapsulating the types of multi-hop questions commonly considered in the question-answering (QA) literature.

We design \ours as a data contamination-resistant tool for controlled stress-testing of LLM reasoning and retrieval capabilities.
(i) On-demand, fully algorithmic generation of previously unseen universes---complete with articles and question-answer pairs---avoids the need of expensive data curation and annotation, encourages the use of fresh dataset instances, and therefore reduces the risk of data leakage.
(ii) Controlling the difficulty and structure of the questions and using logic programming methods to find an exhaustive list of solutions allows for comprehensive testing of both \textit{multi-step} and \textit{multi-branch} reasoning, pinpointing the limits of LLM reasoning capabilities.
(iii) Adjusting the universe size---and therefore the size of the reference document corpus---facilitates testing in a range of LLM settings, such as in-context learning in both short and long context windows, retrieval-augmented generation (RAG), and external tool use (as is common in agentic workflows).

Our evaluation on \ours confirms that the benchmark presents significant challenges for all state-of-the-art LLMs that we used. As a representative example, in \Cref{fig:size-difficulty-f1} we plot F1 scores for \llama for in-context, RAG, and agentic settings at varying reasoning and retrieval difficulties---demonstrating the usefulness of \ours despite the simplicity of its text corpus and otherwise trivially solvable questions.

%% file: sections/related.tex
\section{Related Work}
\label{sec:related_work}

\begin{figure*}[t!]    
\includegraphics[width=\textwidth]{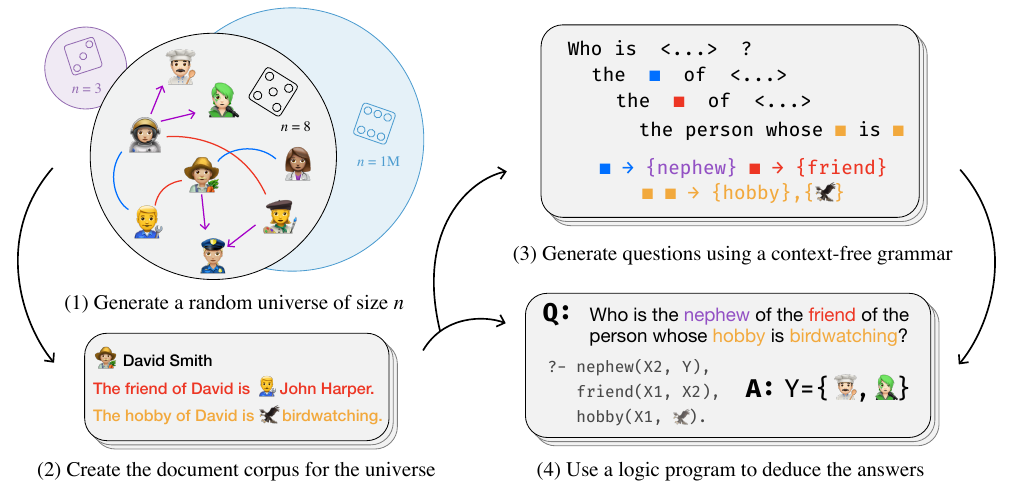}
    \caption{\tbf{Overview of the \ours pipeline.} 
    }
    \label{fig:pipeline}
\end{figure*}

\textbf{Question-Answering Benchmarks.}
A standard technique to evaluate machine reading comprehension is through question-answering benchmarks such as SQuAD \citep{rajpurkar2016squad}, HotpotQA \citep{yang2018hotpotqa}, ComplexWebQuestions \cite{talmor2018web}, QAngaroo \citep{welbl2018constructing}, DROP \citep{dua2019drop}, 2WikiMultiHopQA \citep{ho2020constructing}, HybridQA \citep{chen2020hybridqa}, MuSiQue \citep{trivedi2022musique}, RepLiQA \citep{monteiro2024repliqa} among many others.
These benchmarks are typically curated by crowdsourcing questions based on excerpts from sources like Wikipedia, crowdsourcing the text corpus itself, or leveraging structured knowledge bases like Wikidata to partially automate the question-generation process \citep{agarwal2020knowledge, ye2022generative}. Such benchmark curation is limited as it is time-consuming and expensive, and the released static benchmarks are prone to eventual data contamination and memorization through internet-scraping \citep{trivedi2022musique, monteiro2024repliqa}.
This has been partially addressed by dynamic frameworks such as RealTime QA \citep{kasai2024realtime} and FreshQA \citep{vu2023freshllms}; however, those frameworks are focused on real-time factual knowledge, while \ours's focus is to evaluate knowledge-independent reasoning and retrieval.

\textbf{Retriever- and Tool-Enabled LLM Agents.}
Modern LLM benchmarking places increasing importance on innovations like longer supported context lengths \citep{hsieh2024ruler, wang2024leave, zhang2024infty, an2023eval, bai2023longbench, li2023loogle}, the ability to reference external documents through retrieval-augmented generation (RAG) \citep{lewis2020retrieval, karpukhin2020dense, guu2020retrieval, petroni2020kilt,saad2023ares,jin2024flashrag, hsia2024ragged, mao2024xrag, rau2024bergen, shi2023replug, borgeaud2022improving, tang2024multihop, su2024bright}, and the use of other external tools like search. 
Many of these benchmarks similarly require at least partial manual curation, do not provide a unique corpus, or, in case they are fully synthetic and customizable \citep{hsieh2024ruler}, are limited in their scope.

Many LLM agent benchmarks \citep{yao2024tau, lattimer2024sparse, shridhar2020alfworld, zhou2023webarena} focus on binary-reward tasks (such as booking a flight, making a purchase), tasks that require navigation across multiple pages, or meeting user preferences. 
Closer to our work are tool-augmented question-answering benchmarks. For example, ToolQA \citep{zhuang2023toolqa} introduces a number of tasks in a range of domains and (combinations of) pre-defined tools (e.g. calculators and databases) to evaluate whether a model answers questions using tools or recalls a memorized answer. Compared to this work, \ours is more focused on long-lasting methods to stress-test long-context logical reasoning, uses limited tools, and resists data contamination without requiring dataset curation or manual verification of generated questions.

\textbf{Logical Reasoning Benchmarks.}
Logical reasoning tasks have become central to LLM evaluation and have garnered significant attention in recent time \citep{zhu2023large}. However, many existing benchmarks do not disentangle the evaluation on logical reasoning with other abilities such as natural language inference and commonsense reasoning \citep{sakaguchi2021winogrande, zellers2019hellaswag, sprague2023musr}.
Another line of work focuses on the synthesis of datasets containing 
a variety of logic reasoning tasks \citep{tafjord2020proofwriter, saparov2022language, liu2020logiqa, han2022folio, weston2015towards}. Closer to our work, CLUTRR \citep{sinha2019clutrr} crowd-source short stories about individuals related through a family graph and ask models to determine the relationship between two individuals. We extend this idea to a fully-synthetic on-demand dataset generation pipeline with more diverse question types of controllable difficulty, and enable evaluation in the modern LLM era at long-context and multi-document scale.

%% file: sections/dataset.tex
\section{\ours Construction}
\label{sec:dataset}

\ours is at its core an on-demand random generator of fictional worlds. 
Similarly to the wiki hosting services popular in film, video games, and literature,\footnote{For example, see \href{https://stardewvalley.fandom.com/}{stardewvalley.fandom.com} or \href{https://www.harrypotter.fandom.com/}{harrypotter.fandom.com}.}
we represent these fictional worlds through Wikipedia-like biographical entries about their characters. We then test the model's retrieval skills and its understanding of the fictional world through an accompanying set of automatically generated question-answer pairs.

\subsection{Generating a \ours Universe}
The first stage of the \ours pipeline generates a random universe of $n$ characters as well as the document corpus describing it, as illustrated in \Cref{fig:pipeline}, (1-2).

\tbf{Generating Characters.}
Each character in a \ours universe is described through its \emph{social relationships} and \emph{personal facts} as shown in \Cref{fig:pipeline} (1). 
For the social relationships, we first generate family trees, following the family tree generator of \citet{hohenecker2020ontology}. We iteratively pick a person and generate their parent or child based on various constraints,\footnote{For example, the number of offspring of a person has to be smaller than some threshold, parents of the people at the maximal tree level will not be generated, etc.} until the user-specified universe size of $n$ people is reached.
The user can also specify other hyperparameters like the number of trees, their maximal depth, and the maximal number of offspring for each person.
In addition to the family trees, we generate a friendship graph using the Erdős–Rényi model (making two people friends with some fixed probability, typically controlled by the desired average number of friendships.)

\tbf{Generating Facts.}
Next, we generate personal facts for each person in the \ours universe. 
Names are assigned during the family generation procedure, with the first name sampled based on the character's gender and the surname based on the family tree, resulting in 15M full names in total.\footnote{We use unique names in our experiments, but \ours also supports repeated names.}
We also add dates of birth in a way that is consistent with the existing family relations, and assign each person a job and a hobby that we uniformly sample from over 300 and 600 options respectively.

\tbf{Generating Articles.}
Given all relevant facts for each person, we convert them into articles using pre-defined templates, e.g. ``The job of David is a farmer. The hobby of David is birdwatching.'' (see \Cref{fig:pipeline}, (2)). 
This construction conveys the necessary information while keeping the articles short (about 160 tokens on average) and thus supports a larger effective universe size. The resulting articles are the only component of \ours available to the model during its evaluation (see \Cref{app:example-corpus} for an example of a small \ours corpus).

While it is possible to improve on the minimalistic templated articles using LLM-based rephrasing (see e.g. \citealt{shao2024assisting}), this poses the challenge of guaranteeing factual correctness without additional costs or external supervision. 
This has been supported by our experiments on articles rephrased using \llama (see \Cref{app:article-rephrasing})---while we did not observe any noticeable differences in performance trends, we identified factual errors (hallucinations) in the rephrased articles. Using templates therefore keeps \ours dataset cheaper (no computational or API cost), faster (no latency from LLM queries), and fully factually consistent; however we do see consistency-preserving LLM rephrasing methods as an exciting future direction.

\subsection{Generating Question-Answer Pairs}
\label{sub:qagen}

In the second half of the \ours pipeline, we generate a set of questions with verifiable answers, as shown in \Cref{fig:pipeline}, (3-4).

\tbf{Generating Questions.}
We implement automatic question generation through
a context-free grammar (CFG, \citealt{hopcroft2001introduction}) of \emph{question templates}, which we then use to sample complete questions.
For example, the question template ``\textit{Who is the \emph{$<$relation$>$} of \emph{$<$name$>$}}?'' 
can be used to sample the question
``\textit{Who is the friend of David?}'' (see \Cref{fig:pipeline}, (3)).
The main advantage of using a CFG is that it efficiently and systematically obtains \emph{all} possible compositions of questions for some recursion depth $d$.
For instance, the following subset of our context-free grammar:
\begin{align*}
    S &\rightarrow\; \t{Who is } R \t{?}\\
    R &\rightarrow\; \t{the $<$\textit{relation}$>$ of } R'\\
    R' &\rightarrow\; R \mid\, \t{$<$\textit{name}$>$}
\end{align*}
can lead to questions ranging from ``\textit{Who is the friend of David?}'' to ``\textit{Who is the nephew of the friend of the brother of David?}'' as $d$ increases. In addition to these nested compositions, our CFG also supports questions about personal attributes (e.g. ``\textit{Who is the person whose hobby is birdwatching?}''), aggregation questions (``\textit{How many brothers does David have?}''), and combinations of all three (``\textit{How many friends does the brother of the person whose hobby is birdwatching have?}'')
(See \Cref{app:cfg} for the full CFG.)

\tbf{Generating Answers.}
To ensure that the answers to the sampled questions are verifiably correct, we represent our generated universe in Prolog, a logic programming language \citep{sterling1994art}. 
Each Prolog program consists of a set of facts known about the world such as \texttt{hobby("David", "birdwatching")}, and a set of rules defining how facts are related to each other, such as \texttt{nephew(X, Y) :- sibling(X, A), son(A, Y)}. The Prolog program uses these facts and rules to deduce the exhaustive set of answers to its \emph{queries} (i.e., the CFG-generated questions).
For example, a question 
``\textit{Who is the nephew of the friend of the person whose hobby is birdwatching?}''
corresponds to the three-statement Prolog query
\texttt{?- nephew(X2, Y), friend(X1, X2), hobby(X1, "birdwatching")},
which returns all people satisfying these constraints in the \ours universe (see \Cref{fig:pipeline} (4)).

To construct the Prolog queries automatically, we modify the CFG algorithm to generate both the question and query templates in parallel. We note, however, that the queries are separate from the final \ours corpus and question-answer pairs, and the answers returned by the Prolog program should be held out as part of the evaluation procedure.

\subsection{\ours Complexity}
\label{sec:complexity}
The goal of \ours is to generate memorization-resistant evaluation datasets that are challenging in both reasoning and retrieval aspects.
In this section, we discuss our conceptual and practical design choices that help us achieve this goal.

\tbf{Universe Space Complexity.}
To ensure that our evaluation with \ours is memorization and data leakage-resistant, we first show that the space of possible universes is sufficiently large to generate enough unique instances.
Observe that the number of possible friendship assignments grows at the rate of $\Theta(2^{n^2})$~\citep[Ex.~II.5]{flajolet2009analytic} as the number of individuals $n$ in the universe increases.
Similarly, assuming each individual is assigned one fact from each category (job, hobby, etc.), the number of possible fact assignments grows at the rate $\Theta(c^n)$, where $c$ is the total number of choices across the categories.
\ours thus samples a corpus from $\Theta ( 2^{n^2} c^n )$ possible universes, which leads to diverse datasets optimal for data leakage-resistant evaluation.
We note that as future work \ours could be extended to increase this diversity, e.g. by adding a temporal dimension of events.

\tbf{Reasoning Complexity.}
The CFG enables us to recursively compose templates that lead to complex reasoning questions.
Observe that our CFG in \Cref{app:cfg} produces $\Theta(d)$ question templates as the recursion depth $d$ increases.
Moreover, we can increase the difficulty of each template by increasing the number of \textit{reasoning steps}.
For example, substituting \textit{$<$relation$>$} with \textit{nephew} in a template adds two reasoning steps (\texttt{nephew(X, Y) :- sibling(X, A), son(A, Y)}), since \ours articles only contain immediate family relationships like \emph{sibling} and \emph{son}.
In contrast, substituting \textit{$<$relation$>$} with \textit{second cousin} would lead to five reasoning steps.
As we will show in \Cref{sec:main_results}, \ours questions are sufficiently complex to evaluate reasoning capabilities of state-of-the-art LLMs.
We further note that \ours's CFG can be easily extended to support more question types like comparison and multiple-constraint questions.

\tbf{Retrieval Complexity.}
To assess a model's retrieval capabilities, we increase the universe size $n$ so that the document corpus exceeds the model's context length---this makes a retriever necessary to answer questions correctly.
For state-of-the-art LLMs with a context length of 128K, such as OpenAI's \gpt and Meta's \llama,  this corresponds to \ours universes of $n \gtrapprox 1$K.
This increases to $n \gtrapprox 3$K for Google's \gemini with context length 1M.
Further scaling $n$ leads to further increase in retrieval difficulty.
In \Cref{tab:timings}, we show that \ours is well-suited for generating universes of this size on standard CPU hardware:
generating questions with recursion depth $d=10$ for size $n = 100$K---well beyond any existing LLM's context length---takes just 6 minutes on 8 Intel Cascade Lake CPU cores.
Moreover, we can conveniently generate instances of $n = 1$M, which is on the scale of Wikipedia's corpus of {2 million biographical entries}.\footnote{\href{https://en.wikipedia.org/wiki/Wikipedia:WikiProject_Biography}{https://en.wikipedia.org/wiki/Wikipedia:WikiProject\_Biography}, as of January 30, 2025.}

\begin{table}[h]
    \centering
    \caption{Runtime breakdown of generating a \ours instance for facts, articles and questions for universe sizes $n$.} 
    \label{tab:timings}
\begin{tabular}{ccccc}
\toprule
$n$ & \tbf{Total Runtime} & \tbf{Facts} & \tbf{Articles} & \tbf{Questions} \\
\midrule
$10^2$ & 0.97~s & 0.46~s & 0.07~s & 0.44~s \\
$10^3$ & 2.86~s & 0.90~s & 0.59~s & 1.37~s \\
$10^4$ & 20.91~s & 5.38~s & 5.87~s & 9.66~s \\
$10^5$ & 5.57~m & 0.81~m & 0.97~m & 3.79~m\\
$10^6$ & 3.86~h & 9.47~m & 11.77~m & 3.51~h\\
\bottomrule
\end{tabular}
\end{table}

%% file: sections/main_results.tex
\begin{table*}
\centering
\caption{\tbf{F1 scores (in \%) for various LLMs and prompting techniques.} We report $\text{mean}\pm\text{standard error}$ across 3 dataset generation seeds (except for \gpt with \CoT and \react due to cost constraints), and indicate the highest F1 score for each $n$ in bold.
\Simpleprompting is infeasible for $n=5$K as the corpus cannot be fully included in the context.
}
\begin{tabular}{ccccccc}
\toprule
Universe Size & Model & \multicolumn{2}{c}{In-Context} & \multicolumn{2}{c}{RAG} & Agentic \\
 &  & \zeroshot & \CoT & \zeroshotrag & \cotrag & \react \\
\midrule
\multirow{4}{*}{50} & DeepSeek-R1-32B & \textbf{42.42 ± 1.69} & \textbf{52.42 ± 2.64} & 19.93 ± 0.49 & 21.51 ± 1.31 & 5.47 ± 1.36 \\
& GPT-4o & 27.20 ± 0.76 & 50.66 & 28.05 ± 2.48 & 20.49 ± 1.07 & \textbf{38.70} \\
& Gemini-1.5-Flash & 28.49 ± 1.15 & 34.61 ± 2.41 & \textbf{28.92 ± 2.60} & 20.12 ± 1.69 & 30.92 ± 1.41 \\
& Llama-3.3-70B & 25.64 ± 0.56 & 48.37 ± 1.75 & 25.18 ± 1.91 & \textbf{27.63 ± 2.27} & 35.83 ± 1.00 \\
\midrule
\multirow{4}{*}{500} & DeepSeek-R1-32B & \textbf{18.33 ± 2.33} & 19.65 ± 3.00 & 16.70 ± 0.85 & 17.87 ± 1.32 & 3.57 ± 0.01 \\
& GPT-4o & 16.76 ± 0.87 & \textbf{41.02} & \textbf{22.32 ± 1.99} & 16.39 ± 0.85 & \textbf{37.39} \\
& Gemini-1.5-Flash & 17.39 ± 1.45 & 25.17 ± 1.77 & 21.47 ± 1.44 & 15.09 ± 1.28 & 26.99 ± 1.84 \\
& Llama-3.3-70B & 11.59 ± 1.19 & 25.99 ± 2.09 & 19.45 ± 0.93 & \textbf{21.60 ± 2.06} & 35.56 ± 0.49 \\
\midrule
\multirow{4}{*}{5000} & DeepSeek-R1-32B & 
    \multicolumn{2}{p{3.6cm}}{\multirow{4}{*}{\centering
    \begin{tcolorbox}[height=1.7cm, colframe=black!0, colback=gray!0, boxrule=0.2mm, width=1.1\linewidth,halign=center,valign=center]
    N/A (exceeds \\maximum context) \end{tcolorbox}}} & 15.64 ± 0.88 & 14.81 ± 1.44 & 4.74 ± 0.04 \\
& GPT-4o &&& \textbf{18.13 ± 0.66} & 14.25 ± 1.54 & \textbf{36.85} \\
& Gemini-1.5-Flash &&& 17.94 ± 0.93 & 12.51 ± 0.94 & 23.47 ± 1.53 \\
& Llama-3.3-70B &&& 15.07 ± 0.55 & \textbf{17.89 ± 0.45} & 30.89 ± 2.24 \\
\bottomrule
\end{tabular}
\label{tab:exp__results}
\end{table*}

\section{Experimental Validation}
\label{sec:main_results}

We evaluate reasoning and retrieval capabilities of several frontier LLMs using \ours, by decomposing their performance over questions of varying difficulty and universes of varying sizes.

\subsection{Evaluation Setup}

We generate \ours instances with $n$ ranging from 50 to 10K---a universe size for which the total length of articles exceed context lengths of all existing LLMs.
For evaluation, we only provide the articles to the LLMs, not the Prolog database or the generated graphs.
To ensure that our findings are not tied to any specific \ours instance, we use 3 random dataset seeds for each configuration.
Creating \ours instances with different random seeds leads to entirely different combinations of names, relations, and personal facts.
In each instance, we generate question templates with maximum recursion depth $d=20$, for a total of 50 templates.
We sample 10 questions for each template, yielding a total of 500 questions per \ours instance.
As shown in \Cref{fig:difficulty_distribution,fig:solution_distribution} (\Cref{app:cfg}),
these questions have varying difficulty and number of answers.
Accordingly, we prompt the LLMs to predict all answers as a comma-separated list and measure correctness with the answer-level F1 score.

\subsection{Models and Prompting Techniques}

We test both open- and closed-source LLMs, namely OpenAI's \gpt~\citep{hurst2024gpt}, Google's \gemini~\citep{google2024gemini}, and the instruction-tuned version of Meta's \llama model~\citep{dubey2024llama}.
We also evaluate DeepSeekAI's \deepseek~\citep{guo2025deepseek} distilled with Qwen-2.5-32B~\citep{yang2024qwen2}, which is an open-weights LLM trained on reasoning trace datasets.
We prompt each LLM with the following techniques, broadly grouped in three ways:

\textbf{In-Context Prompting.} This technique includes the full document corpus as part of the prompt. We implement this technique with two strategies: \zeroshot, where the document corpus is immediately followed by the question, and Chain-of-Thought (\CoT)~\citep{wei2022cot}, where we additionally include a few examples of step-by-step reasoning leading to the correct answer. See \Cref{sub:zeroshot_simple} for the prompt details.

\textbf{RAG Prompting.} 
This setting augments generation with a retriever as pioneered by \citet{lewis2020retrieval}. Due to its synthetic nature, the documents from \ours do not necessarily match real-world corpora, making neural retrievers a poor fit for evaluation. Instead, we use the BM25 retriever, which uses keyword matching, and search for the top-4 most relevant documents for each question. Next, we incorporate these retrieved documents into the model's prompt. Finally, we add in the same \zeroshot and \CoT prompts as In-Context Prompting. See \Cref{sub:zeroshot_rag} for details about our retrieval setup.

\textbf{Agentic Prompting.} \react~\citep{yao2022react} is a prompting technique that enables LLMs to interleave reasoning steps with tool interactions, to solve complex tasks. For \ours QA task, the LLMs are provided with keyword-based tools \texttt{RetrieveArticle} and \texttt{Search} to retrieve relevant documents. 
See \Cref{sub:react} for tool details.
These settings materialize the limitations of \simpleprompting and necessitate the use of advanced \ragprompting and \agenticprompting approaches.

In the \CoT and \react prompts, we include 10 QA exemplars and hand-written reasoning traces.
We choose these exemplars from a dataset instance of size 25 that is not used for evaluation.
In \react, we limit LLMs to interact with the text corpus for up to 50 steps, which is sufficient to answer almost all questions in \ours instances.

We cap all LLM outputs to 4096 tokens and use greedy decoding ($\text{temperature}=0$).
For \deepseek, we use $\text{temperature}=0.6$ and $\text{top-$p$}=0.95$, following \citet[Section~3]{guo2025deepseek}.

\subsection{Results}

In \Cref{tab:exp__results}, we report the mean F1 score across various universe sizes, LLMs, and prompting techniques.
We first average F1 scores over all questions in a \ours instance, then compute the mean and standard error across the dataset generation seeds.

We first consider the small-universe setting ($n=50$) in \Cref{tab:exp__results}, which corresponds to roughly 16K tokens for the LLMs we test.
\Simpleprompting techniques outperform other techniques: \CoT with \deepseek attains the highest performance, followed by \CoT with \gpt. 
Next, we consider the setting of medium universes ($n=500$). 
Here the full document corpus can still be included in all LLMs' contexts, but we find that \zeroshot starts to perform poorly for all LLMs, and \deepseek especially struggles.
F1 scores of \CoT for all LLMs degrade as well compared to $n=50$, but perform comparably to \react workflow.
Finally, in the setting of large universes ($n=5000$), none of the LLMs we evaluate can accommodate the full document corpus.
As the in-context techniques are no longer sufficient, we must rely on \ragprompting and \agenticprompting.
\ragprompting attains poor F1 scores because the retriever fails to retrieve documents relevant for answering complex questions.
On the other hand, \agenticprompting technique shines in comparison to other techniques, indicating that LLMs are better suited to dynamically retrieve documents while reasoning on a question.
We attribute the poor performance of \deepseek with \agenticprompting to its inferior tool-calling abilities compared to the other LLMs.

%% file: sections/reasoning.tex
\begin{figure*}[t!]
    \centering
    \includegraphics[width=\linewidth]{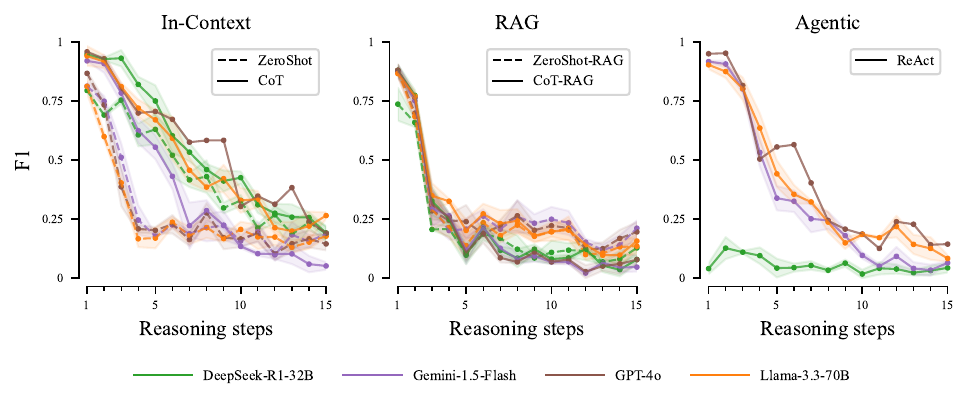}
    \caption{%
    \textbf{
    F1 scores versus question difficulty, measured by \textit{reasoning steps}.
    }%
    We plot LLM performance on universe size $n=50$, and report F1 scores averaged over 3 generation seeds.
    Increasing question difficulty in \ours reveals a clear decline across all state-of-the-art LLMs and prompting techniques, showing their struggle with reasoning.
    }
    \label{fig:exp__f1_v_difficulty}
\end{figure*}

\section{Evaluating Reasoning}
\label{sec:reasoning}

To isolate LLM reasoning capabilities on \ours, we investigate model performance on small universes ($n=50$) in \Cref{fig:exp__f1_v_difficulty}.
Note that the context windows of all LLMs can fully include small universe document corpora.
Each \ours dataset contains questions covering a wide range of difficulty. 
We evaluate three approaches: \simpleprompting, \ragprompting, and \agenticprompting.
For each we plot the F1 scores as a function of question difficulty, as measured by the number of reasoning steps necessary to answer the question.
As mentioned in \Cref{sec:complexity}, this is determined by the type of question templates and the sampled relationships.
For all LLMs and prompting techniques, we verify empirically that \textbf{questions with a  larger number of reasoning steps are indeed more challenging to answer}. 

\zeroshotsimple performance declines sharply as the number of reasoning steps increases for all LLMs, except for \deepseek, which deteriorates more gradually. LLMs perform better with \cotsimple than with \zeroshotsimple, but each additional reasoning step remains increasingly challenging. Another failure mode multi-branch reasoning, i.e. in keeping track of multiple reasoning traces that could lead to valid solutions. For example, a model might fail to find all possible solutions to a question ``Who is the great-grandchild of David?'' by forgetting to check some of the grandchildren; this compounds further as the number of possible constraints increases. This suggests that even in the absence of retrieval-specific constraints, LLMs struggle to navigate logical reasoning sequences.

\ragprompting techniques (\zeroshotrag and \cotrag) stunt reasoning performance across the board---F1 scores are near zero on questions with 5 or more reasoning steps as opposed to 15 steps for \simpleprompting.
We attribute this to a core problem with \ragprompting: retrieving documents in the initial prompt before starting to answer the question, as opposed to reasoning through the question and retrieving documents dynamically.

We find that \ragprompting techniques can only answer questions that require a single reasoning step (e.g., ``Who is the friend of David?'').
Thus, answering questions that require information from \textit{multiple} reasoning steps is extremely challenging for \zeroshotrag and \cotrag. 
Consider the question, ``Who is the nephew of the friend of David?,'' which requires retrieving David's document first, then retrieving their friend's document to find the nephew. Since \ragprompting techniques retrieve documents \textit{only once} by matching vector embeddings of questions and documents, they are unlikely to retrieve all necessary documents required to answer such questions. 
Recent multi-hop RAG prompting methods exhibit slightly better performance on \ours, but similarly struggle on questions requiring many reasoning steps, as shown in \cref{app:multi_hop_rag}.

Finally, the \agenticprompting technique \react allows LLMs to avoid the steep initial performance drop as seen in \ragprompting.
On given a question, \react prompting requires LLMs to retrieve documents dynamically in a conversation and justify why they are relevant.
Concretely, before using a tool (\texttt{RetrieveArticle} or \texttt{Search}) in a conversation turn, the LLM is asked to describe how the tool will help using a ``Thought'' step~\citep{yao2022react}, analogous to the \CoT prompting approach.
This approach shows promise in answering questions correctly.
Even so, \react struggles as the question difficulty increases.

\Cref{fig:exp__f1_v_difficulty} thus decomposes LLM performance along the lines of reasoning capabilities.
It reveals that all \simpleprompting and \agenticprompting achieve near-perfect F1 scores on low-difficulty questions.
Therefore, the stratification between them in \Cref{tab:exp__results} can be attributed to varying performance on high difficulty questions. 
To further isolate the impact of question difficulty, in \Cref{fig:exp__f1_sol1_v_difficulty} we plot F1 scores as a function of reasoning steps for questions with only one solution.

%% file: sections/retrieval.tex
\begin{figure*}[t!]
    \centering
    \includegraphics[width=\linewidth]{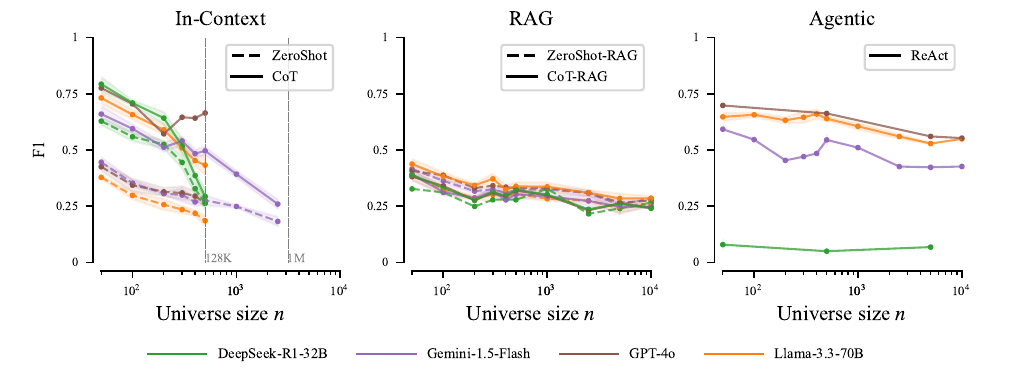}
    \caption{
    \textbf{F1 scores as a function of universe size $n$.}
    We evaluate LLM performance on questions with $\leq 10$ reasoning steps, and report F1 scores averaged over 3 dataset generation seeds.
    As we increase universe size in \ours, F1 scores for all LLMs and prompting techniques deteriorate---albeit at different rates.
    }
    \label{fig:size-f1}
\end{figure*}

\section{Evaluating Retrieval}
\label{sec:retrieval}

Next, to evaluate LLM retrieval capabilities, we use \ours to contrast two settings: (1) small universes where the document corpus can comfortably fit in LLM context, and (2) large universes where the full corpus exceeds context lengths.
To this end, we increase the universe size up to $n =10$K, which corresponds to document corpora well beyond the context lengths of state-of-the-art LLMs, and display the results in \Cref{fig:size-f1}.

For small universes ($n=50$), \CoT outperforms \zeroshot for all LLMs.
However, F1 scores noticeably worsen as more documents are included in models' contexts, with \deepseek suffering a dramatic performance drop.This analysis regime indicates that \textbf{state-of-the-art LLMs struggle at in-context retrieval} for complex question-answering tasks.

At the large universe scale, \simpleprompting techniques become nonviable as the document corpus exceeds model context lengths.
Therefore the use of out-of-context retrieval, such as \ragprompting and \agenticprompting techniques, is necessary for obtaining the answers.
Here we observe that \ragprompting techniques, whose performance is heavily determined by the quality of retrieved documents, deliver sub-par F1 scores across all universe sizes. As expected, the performance declines with increasing universe size.
Interestingly, chain-of-thought does not always improve performance over zeroshot in the RAG setting. In fact, \gemini and \gpt with \cotrag underperforms their \zeroshotrag counterparts.
When chain-of-thought does provide an improvement, however, the gap between CoT and zeroshot is narrower in the RAG setting than in the in-context setting.
Thus, we observe that \zeroshotrag outperforms \zeroshot, but \cotrag underperforms \CoT (assuming the corpus can fit into context).
\Agenticprompting techniques like \react show immense promise by avoiding a steep downward trend.
This suggests that \textbf{agentic workflows can be effective in dynamically retrieving documents at scale}.

%% file: sections/finetuning_on_pw.tex
\section{Fine-tuning on \ours}

\ours generates on-demand datasets with random facts and relationships, ensuring that each dataset instance is unique and evaluation is resistant to fact memorization.
Even so, it is possible that fine-tuning on \ours-generated datasets could improve performance by leveraging linguistic structure in \ours articles and questions.
We fine-tune LLMs on \ours datasets, simultaneously (1) assessing the viability of \ours for training LLMs and (2) testing the robustness of \ours to memorization when evaluating LLMs.

We fine-tune Qwen2.5 0.5B and 3B LLMs~\citep{yang2024qwen2} on 10 new \ours dataset instances with two popular fine-tuning algorithms: group relative policy optimization (GRPO)~\citep{shao2024deepseekmath} and supervised fine-tuning (SFT), and evaluate on the three \ours datasets of size $n=50$ used in \cref{tab:exp__results} (refer to \cref{sub:finetuning_details} for details).
We find that GRPO improves F1 scores over \simpleprompting techniques (\cref{tab:fine_tuning}) and SFT does not.
The performance is still far from optimal, however, and the F1 scores decline as the number of reasoning steps increases (\cref{fig:fine_tuning}), imitating the phenomenon in \cref{fig:exp__f1_v_difficulty}.
This demonstrates that \ours evaluation is robust to LLM memorization, and paves the way for further research in improving LLM reasoning capabilities.

\begin{table}[!h]
    \caption{\textbf{Evaluation F1 scores (in \%) of LLMs fine-tuned on \ours.} As in \cref{tab:exp__results}, we report the mean $\pm$ standard error across 3 dataset generation seeds for universe size $n=50$. \zeroshot and \CoT are prompting methods; SFT and GRPO are fine-tuning methods. Qwen2.5-0.5B \CoT performance is notably worse than \zeroshot since the small model fails to follow the desired answer template.}
    \centering
    \begin{tabular}{ccc}
        \toprule
        \textbf{Method} & \textbf{Qwen2.5-0.5B} & \textbf{Qwen2.5-3B (LoRA)} \\
        \midrule
        \zeroshot & 11.78 $\pm$ 0.94 & 16.82 $\pm$ 2.37 \\
        \CoT & 2.68 $\pm$ 0.22 & 13.71 $\pm$ 0.81 \\
        SFT & 11.71 $\pm$ 1.10 & 16.89 $\pm$ 2.22 \\
        GRPO & 13.25 $\pm$ 0.93 & 31.38 $\pm$ 0.86 \\
        \bottomrule
    \end{tabular}
    \label{tab:fine_tuning}
\end{table}

%% file: sections/future.tex
\section{Conclusion and Future Work}
\label{sec:future}

We introduce \ours---a framework for benchmarking the reasoning and retrieval capabilities of language models.
As we increase the question complexity and universe size, we observe that current state-of-the-art LLMs struggle in terms of both reasoning and retrieval.
\ours is scalable and memorization-resistant, hence well-suited to evaluate future generations of language models.

Our work brings forth several research directions. One of them could be to extend \ours from its limited setting and form to support more complex types of facts and relations. The simplicity and minimalism of the text corpus could be improved using LLM-based paraphrasing methods, which, as noted before, would require innovation in preserving consistency of and preventing hallucinations in the data. 
In this work we focus on question-answering over text corpora;
this leaves potential to extend \ours to other knowledge bases and modalities such as vision and audio, enabling analogous test suites in multimodal settings.

%% file: appendix.tex
\newpage
\appendix
\onecolumn
\input{appendices/background}
\input{appendices/CFG}
\input{appendices/baselines}
\input{appendices/additional_results}
\input{appendices/small_corpus}

%% file: appendices/background.tex
\section{Background}
\subsection{Context-Free Grammars}

Context-free grammar (CFG) is a type of formal grammar where the productions rules govern how to generate text from non-terminals and terminals.
A context-free grammar is defined by $G=(V,\Sigma, R, S)$ where $V$ and $\Sigma$ denotes nonterminal and terminal respectively. $R$ is a finite relation in $V \times (V \cup \Sigma)^{*}$ which specifies the production rules of the grammar. $S \in V$ is the start symbol. 
A production rule in $R$ has the form 
\begin{equation}
    \alpha \to \beta
\end{equation}
where $\alpha \in V$, $\beta \in (V \cup \Sigma)^{*}$.
It is conventional to list all rules with the same left-hand side on the same line and separate the right-hand side with ``$|$'' like $\alpha \to \beta_1 \,|\, \beta_2$.

%% file: appendices/CFG.tex
\section{Question Template Generation}
\label{app:cfg}

\subsection{Context-Free Grammar}
\label{app:question_cfg}

We use the following CFG to generate question templates:
\begin{verbatim}  
    S -> Who is R? | What is A ? | How many RN_p does R_c have ?
    R -> the RN of R_c | the person whose AN is AV
    R_c -> R | N
    A -> the AN of R
    RN -> <relation>
    RN_p -> <relation_plural>
    AN -> <attribute_name>
    AV -> <attribute_value>
    N -> <name>
\end{verbatim}

\subsection{CFG-generated question templates}
\label{app:question_templates}

Our CFG produces the following 50 question templates at recursion depth $d = 20$.
Note how the recursive production rule \verb+R_c -> R | N+ leads to chained productions.

\sloppy
\begin{lstlisting}[basicstyle=\ttfamily,breaklines=true]
1. Who is the <relation>_3 of the <relation>_5 of the <relation>_7 of the <relation>_9 of the <relation>_11 of the <relation>_13 of the <relation>_15 of the <relation>_17 of the person whose <attribute_name>_19 is <attribute_value>_19?
2. Who is the <relation>_3 of the <relation>_5 of the <relation>_7 of the <relation>_9 of the <relation>_11 of the <relation>_13 of the <relation>_15 of the <relation>_17 of <name>_18?
3. Who is the <relation>_3 of the <relation>_5 of the <relation>_7 of the <relation>_9 of the <relation>_11 of the <relation>_13 of the <relation>_15 of the person whose <attribute_name>_17 is <attribute_value>_17?
4. Who is the <relation>_3 of the <relation>_5 of the <relation>_7 of the <relation>_9 of the <relation>_11 of the <relation>_13 of the <relation>_15 of <name>_16?
5. Who is the <relation>_3 of the <relation>_5 of the <relation>_7 of the <relation>_9 of the <relation>_11 of the <relation>_13 of the person whose <attribute_name>_15 is <attribute_value>_15?
6. Who is the <relation>_3 of the <relation>_5 of the <relation>_7 of the <relation>_9 of the <relation>_11 of the <relation>_13 of <name>_14?
7. Who is the <relation>_3 of the <relation>_5 of the <relation>_7 of the <relation>_9 of the <relation>_11 of the person whose <attribute_name>_13 is <attribute_value>_13?
8. Who is the <relation>_3 of the <relation>_5 of the <relation>_7 of the <relation>_9 of the <relation>_11 of <name>_12?
9. Who is the <relation>_3 of the <relation>_5 of the <relation>_7 of the <relation>_9 of the person whose <attribute_name>_11 is <attribute_value>_11?
10. Who is the <relation>_3 of the <relation>_5 of the <relation>_7 of the <relation>_9 of <name>_10?
11. Who is the <relation>_3 of the <relation>_5 of the <relation>_7 of the person whose <attribute_name>_9 is <attribute_value>_9?
12. Who is the <relation>_3 of the <relation>_5 of the <relation>_7 of <name>_8?
13. Who is the <relation>_3 of the <relation>_5 of the person whose <attribute_name>_7 is <attribute_value>_7?
14. Who is the <relation>_3 of the <relation>_5 of <name>_6?
15. Who is the <relation>_3 of the person whose <attribute_name>_5 is <attribute_value>_5?
16. Who is the <relation>_3 of <name>_4?
17. Who is the person whose <attribute_name>_3 is <attribute_value>_3?
18. What is the <attribute_name>_3 of the <relation>_4 of the <relation>_6 of the <relation>_8 of the <relation>_10 of the <relation>_12 of the <relation>_14 of the <relation>_16 of the <relation>_18 of <name>_19?
19. What is the <attribute_name>_3 of the <relation>_4 of the <relation>_6 of the <relation>_8 of the <relation>_10 of the <relation>_12 of the <relation>_14 of the <relation>_16 of the person whose <attribute_name>_18 is <attribute_value>_18?
20. What is the <attribute_name>_3 of the <relation>_4 of the <relation>_6 of the <relation>_8 of the <relation>_10 of the <relation>_12 of the <relation>_14 of the <relation>_16 of <name>_17?
21. What is the <attribute_name>_3 of the <relation>_4 of the <relation>_6 of the <relation>_8 of the <relation>_10 of the <relation>_12 of the <relation>_14 of the person whose <attribute_name>_16 is <attribute_value>_16?
22. What is the <attribute_name>_3 of the <relation>_4 of the <relation>_6 of the <relation>_8 of the <relation>_10 of the <relation>_12 of the <relation>_14 of <name>_15?
23. What is the <attribute_name>_3 of the <relation>_4 of the <relation>_6 of the <relation>_8 of the <relation>_10 of the <relation>_12 of the person whose <attribute_name>_14 is <attribute_value>_14?
24. What is the <attribute_name>_3 of the <relation>_4 of the <relation>_6 of the <relation>_8 of the <relation>_10 of the <relation>_12 of <name>_13?
25. What is the <attribute_name>_3 of the <relation>_4 of the <relation>_6 of the <relation>_8 of the <relation>_10 of the person whose <attribute_name>_12 is <attribute_value>_12?
26. What is the <attribute_name>_3 of the <relation>_4 of the <relation>_6 of the <relation>_8 of the <relation>_10 of <name>_11?
27. What is the <attribute_name>_3 of the <relation>_4 of the <relation>_6 of the <relation>_8 of the person whose <attribute_name>_10 is <attribute_value>_10?
28. What is the <attribute_name>_3 of the <relation>_4 of the <relation>_6 of the <relation>_8 of <name>_9?
29. What is the <attribute_name>_3 of the <relation>_4 of the <relation>_6 of the person whose <attribute_name>_8 is <attribute_value>_8?
30. What is the <attribute_name>_3 of the <relation>_4 of the <relation>_6 of <name>_7?
31. What is the <attribute_name>_3 of the <relation>_4 of the person whose <attribute_name>_6 is <attribute_value>_6?
32. What is the <attribute_name>_3 of the <relation>_4 of <name>_5?
33. What is the <attribute_name>_3 of the person whose <attribute_name>_4 is <attribute_value>_4?
34. How many <relation_plural>_2 does the <relation>_4 of the <relation>_6 of the <relation>_8 of the <relation>_10 of the <relation>_12 of the <relation>_14 of the <relation>_16 of the <relation>_18 of <name>_19 have?
35. How many <relation_plural>_2 does the <relation>_4 of the <relation>_6 of the <relation>_8 of the <relation>_10 of the <relation>_12 of the <relation>_14 of the <relation>_16 of the person whose <attribute_name>_18 is <attribute_value>_18 have?
36. How many <relation_plural>_2 does the <relation>_4 of the <relation>_6 of the <relation>_8 of the <relation>_10 of the <relation>_12 of the <relation>_14 of the <relation>_16 of <name>_17 have?
37. How many <relation_plural>_2 does the <relation>_4 of the <relation>_6 of the <relation>_8 of the <relation>_10 of the <relation>_12 of the <relation>_14 of the person whose <attribute_name>_16 is <attribute_value>_16 have?
38. How many <relation_plural>_2 does the <relation>_4 of the <relation>_6 of the <relation>_8 of the <relation>_10 of the <relation>_12 of the <relation>_14 of <name>_15 have?
39. How many <relation_plural>_2 does the <relation>_4 of the <relation>_6 of the <relation>_8 of the <relation>_10 of the <relation>_12 of the person whose <attribute_name>_14 is <attribute_value>_14 have?
40. How many <relation_plural>_2 does the <relation>_4 of the <relation>_6 of the <relation>_8 of the <relation>_10 of the <relation>_12 of <name>_13 have?
41. How many <relation_plural>_2 does the <relation>_4 of the <relation>_6 of the <relation>_8 of the <relation>_10 of the person whose <attribute_name>_12 is <attribute_value>_12 have?
42. How many <relation_plural>_2 does the <relation>_4 of the <relation>_6 of the <relation>_8 of the <relation>_10 of <name>_11 have?
43. How many <relation_plural>_2 does the <relation>_4 of the <relation>_6 of the <relation>_8 of the person whose <attribute_name>_10 is <attribute_value>_10 have?
44. How many <relation_plural>_2 does the <relation>_4 of the <relation>_6 of the <relation>_8 of <name>_9 have?
45. How many <relation_plural>_2 does the <relation>_4 of the <relation>_6 of the person whose <attribute_name>_8 is <attribute_value>_8 have?
46. How many <relation_plural>_2 does the <relation>_4 of the <relation>_6 of <name>_7 have?
47. How many <relation_plural>_2 does the <relation>_4 of the person whose <attribute_name>_6 is <attribute_value>_6 have?
48. How many <relation_plural>_2 does the <relation>_4 of <name>_5 have?
49. How many <relation_plural>_2 does the person whose <attribute_name>_4 is <attribute_value>_4 have?
50. How many <relation_plural>_2 does <name>_3 have?
\end{lstlisting}

\subsection{Question-Answer Characteristics}

\begin{figure}[h]
    \centering
    \includegraphics[width=0.3\linewidth]{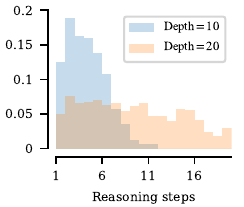}
    \caption{Histogram of question difficulties (measured by reasoning steps) for universe size $n = 50$ at two CFG recursion depths $d \in \set{10, 20}$.
    We average the frequencies across 3 dataset generation seeds.}
    \label{fig:difficulty_distribution}
\end{figure}

\begin{figure}[h]
    \centering
    \includegraphics[width=0.5\linewidth]{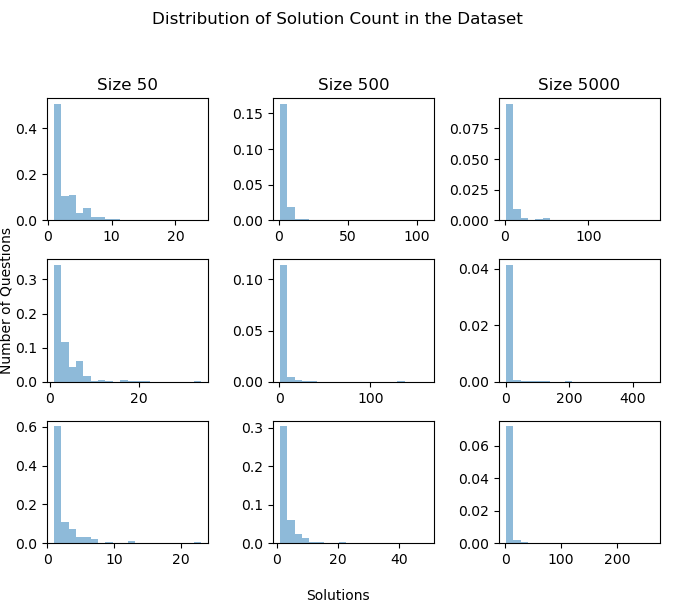}
    \caption{Distribution of number of answers across sizes $n \in \braces{50,500,5000}$, seeds $\braces{1,2,3}$, and CFG depth 20.}
    \label{fig:solution_distribution}
\end{figure}

%% file: appendices/baselines.tex
\section{Baseline Details}\label{app:baselines}

\subsection{\textsc{ZeroShot-Simple}} \label{sub:zeroshot_simple}

We use the following prompt for all models, where \verb+evidence+ is the concatenation of all documents in the \ours instance. 

\begin{lstlisting}[basicstyle=\ttfamily,breaklines=true]
You are given the following evidence:
(BEGIN EVIDENCE)
{{evidence}}
(END EVIDENCE)

You will be provided a question. Your task is to provide an answer according to these instructions: 
- The output must be one of the following: a name (if there is only one correct answer); or a list of names separated by '{constants.answer_sep}' (if there are multiple correct answers).
- DO NOT include any additional information in your answer.

Question: {{question}}
Answer: 
\end{lstlisting}

For \deepseek, we additionally parse the output to separate the model's reasoning process from its final answer using the \texttt{$<$/think$>$} tag.

\subsection{\textsc{ZeroShot-RAG}} \label{sub:zeroshot_rag}

The prompt is exactly the same as \zeroshotsimple, except we replace \verb+evidence+ with 4 documents retrieved using the BM25 retriever.
Upon generation, we search for similar documents for \verb+question+ according to maximum inner product search on document and question embeddings.

\subsection{\textsc{Chain-of-Thought-Simple}} \label{sub:cot_simple}

We use the following prompt for all models, where \verb+evidence+ is replaced with a list of all documents.
We use a regular expression to parse the output.

\begin{lstlisting}[basicstyle=\ttfamily,breaklines=true]
You are given the following evidence:
(BEGIN EVIDENCE)
{{evidence}}
(END EVIDENCE)

You will be provided a question. Your response must end in the following sentence: The answer is <answer>.
Here, <answer> must be one of the following: 
- a name (if there is only one correct answer); or
- a list of names separated by '{constants.answer_sep}' (if there are multiple correct answers).

Here are some examples:
(START OF EXAMPLES)
Example 1:
Question: Who is the brother of Dino Beltran?
Answer: Based on the evidence, the brother of Dino Beltran is Orlando Beltran. The answer is Orlando Beltran.

Example 2:
Question: Who is the sibling of Barabara Beltran?
Answer: Based on the evidence, the siblings of Barabara Beltran are Aida Wang, Vicki Hackworth. The answer is Aida Wang{constants.answer_sep}Vicki Hackworth.

Example 3:
Question: Who is the child of the sibling of Stacia Toombs?
Answer: First I need to find the sibling of Stacia Toombs. Based on the evidence, the sibling of Stacia Toombs is Shelli Beltran. Now I need to find the child of Shelli Beltran. Based on the evidence, the children of Shelli Beltran are Aida Wang, Barabara Beltran, Vicki Hackworth. The answer is Aida Wang{constants.answer_sep}Barabara Beltran{constants.answer_sep}Vicki Hackworth.

Example 4:
Question: Who is the uncle of William Smock?
Answer: An uncle is the brother of a parent. Based on the evidence, the parents of William Smock are Dominique Smock, Gene Smock. To find the uncle of William Smock, I need to find the brother of Dominique Smock and Gene Smock. Based on the evidence, Dominique Smock has no brother, and the brother of Gene Smock is Eli Smock. So the uncle of William Smock is Eli Smock. The answer is Eli Smock.

Example 5:
Question: What is the occupation of the sister of the grandmother of Virgil Hackworth?
Answer: A grandmother is the mother of a parent. Based on the evidence, the parents of Virgil Hackworth are Ricardo Hackworth, Vicki Hackworth. To find the grandmother of Virgil Hackworth, I need to find the mother of Ricardo Hackworth and Vicki Hackworth. Based on the evidence, Ricardo Hackworth has no mother, and the mother of Vicki Hackworth is Shelli Beltran. Now I need to find the sister of Shelli Beltran. Based on the evidence, the sister of Shelli Beltran is Stacia Toombs. Based on the evidence, the occupation of Stacia Toombs is actuary. The answer is actuary.

Example 6:
Question: Who is the brother of the person whose occupation is associate professor?
Answer: I need to search for people whose occupation is associate professor. Based on the evidence, the person whose occupation is associate professor is Dino Beltran. And the brother of Dino Beltran is Orlando Beltran. The answer is Orlando Beltran.

Example 7:
Question: What is the date of birth of the person whose hobby is meteorology?
Answer: I need to search for people whose hobby is meteorology. Based on the evidence, the people whose hobby is meteorology are Alison Smock, Barabara Beltran. The date of birth of Alison Smock is 0929-10-28, and the date of birth of Barabara Beltran is 0989-06-11. The answer is 0929-10-28{constants.answer_sep}0989-06-11.

Example 8:
Question: Who is the cousin of the person whose occupation is broadcast engineer?
Answer: I need to search for people whose occupation is broadcast engineer. Based on the evidence, the person whose occupation is broadcast engineer is Barabara Beltran. A cousin is the child of the sibling of the parent. Based on the evidence, the parents of Barabara Beltran are Dino Beltran, Shelli Beltran. The sibling of Dino Beltran is Orlando Beltran, and the sibling of Shelli Beltran is Stacia Toombs. Based on the evidence, Orlando Beltran has no child, and the child of Stacia Toombs is Leslee Toombs. So the cousin of Barabara Beltran is Leslee Toombs. The answer is Leslee Toombs.

Example 9:
Question: Who is the great-granddaughter of the person whose hobby is biology?
Answer: I need to search for people whose hobby is biology. Based on the evidence, the person whose hobby is biology is Alvaro Smock. To find the great-granddaughter of Alvaro Smock, I need to find the daughter of the child of the child of Alvaro Smock. Based on the evidence, the children of Alvaro Smock are Eli Smock, Gene Smock. Eli Smock has no child, and the child of Gene Smock is Williams Smock. The daughters of Williams Smock are Shelli Beltran, Stacia Toombs. So the great-granddaughters of Alvaro Smock, whose hobby is biology, are Shelli Beltran, Stacia Toombs. The answer is Shelli Beltran{constants.answer_sep}Stacia Toombs.
(END OF EXAMPLES)

Question: {{question}}
Answer: 
\end{lstlisting}

\subsection{\textsc{Chain-of-Thought-RAG}} \label{sub:cot_rag}

The prompt is exactly the same as \cotsimple, except we replace \verb+evidence+ with 4 documents retrieved using BM25.
We use the same retriever setup as described in \Cref{sub:zeroshot_rag}.

\subsection{\textsc{ReAct}} \label{sub:react}

We modify the ReAct agent implementation of \citet{shinn2024reflexion}, which can be found at \url{https://github.com/noahshinn/reflexion}, to support our keyword-based search tools \texttt{RetrieveArticle} and \texttt{Search}.
As seen in the following examples, the \texttt{RetrieveArticle} tool fetches the documents by title, and the \texttt{Search} tool fetches all document titles whose bodies contain the argument.
We instruct the LLM to output predict the answer using a third tool \texttt{Finish}.

\begin{lstlisting}[basicstyle=\ttfamily,breaklines=true]
Solve a question answering task with interleaving Thought, Action, Observation steps.
Thought can reason about the current situation, and Action can be 3 types:
(1) RetrieveArticle[{{{{entity}}}}]. This action retrieves the article about {{{{entity}}}} if it exists. If the article does not exist, the action will say so.
(2) Search[{{{{attribute}}}}]. This action searches the database for {{{{attribute}}}} and retrieves all articles that contain {{{{attribute}}}}. If no article contains {{{{attribute}}}}, the action will say so.
(3) Finish[{{{{answer}}}}]. This action answers the question with {{{{answer}}}}.
If you cannot find the answer, output the empty answer like: Finish[].
If there are multiple answers A,B,C, answer with a list like: Finish[A{constants.answer_sep}B{constants.answer_sep}C].

You may take as many steps as necessary.
Here are some examples:
(START OF EXAMPLES)
Example 1:
Question: Who is the brother of Dino Beltran?
Thought 1: I need to retrieve article about Dino Beltran and find who his brother is.
Action 1: RetrieveArticle[Dino Beltran].
Observation 1: # Dino Beltran ## Family The parents of Dino Beltran are Brian Beltran, Daisy Beltran.  Dino Beltran's sibling is Orlando Beltran.  The brother of Dino Beltran is Orlando Beltran.  The mother of Dino Beltran is Daisy Beltran.  The father of Dino Beltran is Brian Beltran.  The children of Dino Beltran are Aida Wang, Barabara Beltran, Vicki Hackworth.  The daughters of Dino Beltran are Aida Wang, Barabara Beltran, Vicki Hackworth.  The wife of Dino Beltran is Shelli Beltran.  ## Friends The friend of Dino Beltran is Alvaro Smock.  ## Attributes The date of birth of Dino Beltran is 0958-08-09.  The occupation of Dino Beltran is associate professor.  The hobby of Dino Beltran is shogi.
Thought 2: The brother of Dino Beltran is Orlando Beltran, so the answer is Orlando Beltran.
Action 2: Finish[Orlando Beltran].

Example 2:
Question: Who is the mother of Ivana Smith?
Thought 1: I need to retrieve article about Ivana Smith and find who her mother is.
Action 1: RetrieveArticle[Ivana Smith].
Observation 1: No article exists for the requested entity. Please try retrieving article for another entity.
Thought 2: Article about Ivana Smith does not exist, so I cannot find the mother of Ivana Smith. I will output empty answer.
Action 2: Finish[].

Example 3:
Question: Who is the sibling of Barabara Beltran?
Thought 1: I need to retrieve article about Barabara Beltran and find who her sibling is.
Action 1: RetrieveArticle[Barabara Beltran].
Observation 1: # Barabara Beltran ## Family The parents of Barabara Beltran are Dino Beltran, Shelli Beltran.  Barabara Beltran's siblings are Aida Wang, Vicki Hackworth.  The sisters of Barabara Beltran are Aida Wang, Vicki Hackworth.  The mother of Barabara Beltran is Shelli Beltran.  The father of Barabara Beltran is Dino Beltran.  ## Friends ## Attributes The date of birth of Barabara Beltran is 0989-06-11.  The occupation of Barabara Beltran is broadcast engineer.  The hobby of Barabara Beltran is meteorology.
Thought 2: The siblings of Barabara Beltran are Aida Wang, Vicki Hackworth. So the answer is Aida Wang, Vicki Hackworth.
Action 2: Finish[Aida Wang{constants.answer_sep}Vicki Hackworth].

Example 4:
Question: Who is the child of the sibling of Stacia Toombs?
Thought 1: First, I need to retrieve article about Stacia Toombs and find who her sibling is.
Action 1: RetrieveArticle[Stacia Toombs].
Observation 1: # Stacia Toombs ## Family The parents of Stacia Toombs are Alison Smock, Williams Smock.  Stacia Toombs's sibling is Shelli Beltran.  The sister of Stacia Toombs is Shelli Beltran.  The mother of Stacia Toombs is Alison Smock.  The father of Stacia Toombs is Williams Smock.  The child of Stacia Toombs is Leslee Toombs.  The daughter of Stacia Toombs is Leslee Toombs.  The husband of Stacia Toombs is Wilbert Toombs.  ## Friends The friends of Stacia Toombs are Brian Beltran, Isiah Lutz, Leeann Hackworth, Lesley Lutz, Ryan Wang.  ## Attributes The date of birth of Stacia Toombs is 0959-03-22.  The occupation of Stacia Toombs is actuary.  The hobby of Stacia Toombs is finance.
Thought 2: The sibling of Stacia Toombs is Shelli Beltran. Now I need to retrieve article about Shelli Beltran and find who her child is.
Action 2: RetrieveArticle[Shelli Beltran].
Observation 2: # Shelli Beltran ## Family The parents of Shelli Beltran are Alison Smock, Williams Smock.  Shelli Beltran's sibling is Stacia Toombs.  The sister of Shelli Beltran is Stacia Toombs.  The mother of Shelli Beltran is Alison Smock.  The father of Shelli Beltran is Williams Smock.  The children of Shelli Beltran are Aida Wang, Barabara Beltran, Vicki Hackworth.  The daughters of Shelli Beltran are Aida Wang, Barabara Beltran, Vicki Hackworth.  The husband of Shelli Beltran is Dino Beltran.  ## Friends The friends of Shelli Beltran are Brian Beltran, Eli Smock, Isiah Lutz, Leslee Toombs, Lesley Lutz, Ryan Wang.  ## Attributes The date of birth of Shelli Beltran is 0958-03-08.  The occupation of Shelli Beltran is occupational therapist.  The hobby of Shelli Beltran is sociology.
Thought 3: The child of Shelli Beltran is Aida Wang, Barabara Beltran, Vicki Hackworth. So the answer is Aida Wang, Barabara Beltran, Vicki Hackworth.
Action 3: Finish[Aida Wang{constants.answer_sep}Barabara Beltran{constants.answer_sep}Vicki Hackworth].

Example 5:
Question: Who is the uncle of William Smock?
Thought 1: First, I need to retrieve article about William Smock. To find the uncle of William Smock, I need to find the brother of his parent.
Action 1: RetrieveArticle[William Smock].
Observation 1: # Williams Smock ## Family The parents of Williams Smock are Dominique Smock, Gene Smock.  The mother of Williams Smock is Dominique Smock.  The father of Williams Smock is Gene Smock.  The children of Williams Smock are Shelli Beltran, Stacia Toombs.  The daughters of Williams Smock are Shelli Beltran, Stacia Toombs.  The wife of Williams Smock is Alison Smock.  ## Friends The friend of Williams Smock is Lannie Smock.  ## Attributes The date of birth of Williams Smock is 0926-04-04.  The occupation of Williams Smock is clinical biochemist.  The hobby of Williams Smock is social studies.
Thought 2: The parents of William Smock are Dominique Smock, Gene Smock. To find the uncle of William Smock, I need to find the brother of Dominique Smock and Gene Smock.
Action 2: RetrieveArticle[Dominique Smock].
Observation 2: # Dominique Smock ## Family The child of Dominique Smock is Williams Smock.  The son of Dominique Smock is Williams Smock.  The husband of Dominique Smock is Gene Smock.  ## Friends The friends of Dominique Smock are Gene Smock, Isiah Lutz, Orlando Beltran, Vicki Hackworth.  ## Attributes The date of birth of Dominique Smock is 0897-09-08.  The occupation of Dominique Smock is sports therapist.  The hobby of Dominique Smock is dominoes.
Thought 3: Dominique Smock has no brother. Now I need to retrieve article about Gene Smock and find his brother.
Action 3: RetrieveArticle[Gene Smock].
Observation 3: # Gene Smock ## Family The parents of Gene Smock are Alvaro Smock, Lannie Smock.  Gene Smock's sibling is Eli Smock.  The brother of Gene Smock is Eli Smock.  The mother of Gene Smock is Lannie Smock.  The father of Gene Smock is Alvaro Smock.  The child of Gene Smock is Williams Smock.  The son of Gene Smock is Williams Smock.  The wife of Gene Smock is Dominique Smock.  ## Friends The friends of Gene Smock are Leeann Hackworth, Leisa Lutz, Ricardo Hackworth, Alvaro Smock, Dominique Smock.  ## Attributes The date of birth of Gene Smock is 0898-08-16.  The occupation of Gene Smock is immunologist.  The hobby of Gene Smock is architecture.
Thought 4: The brother of Gene Smock is Eli Smock, and Dominique Smock has no brother. Since Gene Smock and Dominique Smock are the parents of William Smock, the uncle of William Smock is Eli Smock. So the answer is Eli Smock.
Action 4: Finish[Eli Smock].

Example 6:
Question: What is the occupation of the sister of the grandmother of Virgil Hackworth?
Thought 1: First, I need to retrieve article about Virgil Hackworth. To find the grandmother of Virgil Hackworth, I need to find the mother of his parent.
Action 1: RetrieveArticle[Virgil Hackworth].
Observation 1: # Virgil Hackworth ## Family The parents of Virgil Hackworth are Ricardo Hackworth, Vicki Hackworth.  Virgil Hackworth's siblings are Leeann Hackworth, Leisa Lutz.  The sisters of Virgil Hackworth are Leeann Hackworth, Leisa Lutz.  The mother of Virgil Hackworth is Vicki Hackworth.  The father of Virgil Hackworth is Ricardo Hackworth.  ## Friends The friends of Virgil Hackworth are Alison Smock, Eli Smock, Orlando Beltran, Ryan Wang.  ## Attributes The date of birth of Virgil Hackworth is 1009-06-23.  The occupation of Virgil Hackworth is sound technician.  The hobby of Virgil Hackworth is wikipedia editing.
Thought 2: The parents of Virgil Hackworth are Ricardo Hackworth, Vicki Hackworth. To find the grandmother of Virgil Hackworth, I need to find the mother of Ricardo Hackworth and Vicki Hackworth.
Action 2: RetrieveArticle[Ricardo Hackworth].
Observation 2: # Ricardo Hackworth ## Family The children of Ricardo Hackworth are Leeann Hackworth, Leisa Lutz, Virgil Hackworth.  The son of Ricardo Hackworth is Virgil Hackworth.  The daughters of Ricardo Hackworth are Leeann Hackworth, Leisa Lutz.  The wife of Ricardo Hackworth is Vicki Hackworth.  ## Friends The friends of Ricardo Hackworth are Gene Smock, Isiah Lutz, Johnetta Wang.  ## Attributes The date of birth of Ricardo Hackworth is 0983-02-24.  The occupation of Ricardo Hackworth is clinical research associate.  The hobby of Ricardo Hackworth is dairy farming.
Thought 3: Ricardo Hackworth has no mother. So I need to retrieve article about Vicki Hackworth and find her mother.
Action 3: RetrieveArticle[Vicki Hackworth].
Observation 3: # Vicki Hackworth ## Family The parents of Vicki Hackworth are Dino Beltran, Shelli Beltran.  Vicki Hackworth's siblings are Aida Wang, Barabara Beltran.  The sisters of Vicki Hackworth are Aida Wang, Barabara Beltran.  The mother of Vicki Hackworth is Shelli Beltran.  The father of Vicki Hackworth is Dino Beltran.  The children of Vicki Hackworth are Leeann Hackworth, Leisa Lutz, Virgil Hackworth.  The son of Vicki Hackworth is Virgil Hackworth.  The daughters of Vicki Hackworth are Leeann Hackworth, Leisa Lutz.  The husband of Vicki Hackworth is Ricardo Hackworth.  ## Friends The friends of Vicki Hackworth are Brian Beltran, Dominique Smock, Eli Smock.  ## Attributes The date of birth of Vicki Hackworth is 0985-05-30.  The occupation of Vicki Hackworth is police officer.  The hobby of Vicki Hackworth is meditation.
Thought 4: The mother of Vicki Hackworth is Shelli Beltran. So the only grandmother of Virgil Hackworth is Shelli Beltran. Now I need to retrieve article about Shelli Beltran and find her sister.
Action 4: RetrieveArticle[Shelli Beltran].
Observation 4: # Shelli Beltran ## Family The parents of Shelli Beltran are Alison Smock, Williams Smock.  Shelli Beltran's sibling is Stacia Toombs.  The sister of Shelli Beltran is Stacia Toombs.  The mother of Shelli Beltran is Alison Smock.  The father of Shelli Beltran is Williams Smock.  The children of Shelli Beltran are Aida Wang, Barabara Beltran, Vicki Hackworth.  The daughters of Shelli Beltran are Aida Wang, Barabara Beltran, Vicki Hackworth.  The husband of Shelli Beltran is Dino Beltran.  ## Friends The friends of Shelli Beltran are Brian Beltran, Eli Smock, Isiah Lutz, Leslee Toombs, Lesley Lutz, Ryan Wang.  ## Attributes The date of birth of Shelli Beltran is 0958-03-08.  The occupation of Shelli Beltran is occupational therapist.  The hobby of Shelli Beltran is sociology.
Thought 5: The sister of Shelli Beltran is Stacia Toombs. Now I need to retrieve article about Stacia Toombs and find her occupation.
Action 5: RetrieveArticle[Stacia Toombs].
Observation 5: # Stacia Toombs ## Family The parents of Stacia Toombs are Alison Smock, Williams Smock.  Stacia Toombs's sibling is Shelli Beltran.  The sister of Stacia Toombs is Shelli Beltran.  The mother of Stacia Toombs is Alison Smock.  The father of Stacia Toombs is Williams Smock.  The child of Stacia Toombs is Leslee Toombs.  The daughter of Stacia Toombs is Leslee Toombs.  The husband of Stacia Toombs is Wilbert Toombs.  ## Friends The friends of Stacia Toombs are Brian Beltran, Isiah Lutz, Leeann Hackworth, Lesley Lutz, Ryan Wang.  ## Attributes The date of birth of Stacia Toombs is 0959-03-22.  The occupation of Stacia Toombs is actuary.  The hobby of Stacia Toombs is finance.
Thought 6: The occupation of Stacia Toombs is actuary. So the answer is actuary.
Action 6: Finish[actuary].

Example 7:
Question: Who is the brother of the person whose occupation is associate professor?
Thought 1: First, I need to search for people whose occupation is associate professor.
Action 1: Search[associate professor].
Observation 1: (1) Dino Beltran
Thought 2: Person whose hobby is associate professor is Dino Beltran. Now I need to retrieve article about Dino Beltran and find his brother.
Action 2: RetrieveArticle[Dino Beltran].
Observation 2: # Dino Beltran ## Family The parents of Dino Beltran are Brian Beltran, Daisy Beltran.  Dino Beltran's sibling is Orlando Beltran.  The brother of Dino Beltran is Orlando Beltran.  The mother of Dino Beltran is Daisy Beltran.  The father of Dino Beltran is Brian Beltran.  The children of Dino Beltran are Aida Wang, Barabara Beltran, Vicki Hackworth.  The daughters of Dino Beltran are Aida Wang, Barabara Beltran, Vicki Hackworth.  The wife of Dino Beltran is Shelli Beltran.  ## Friends The friend of Dino Beltran is Alvaro Smock.  ## Attributes The date of birth of Dino Beltran is 0958-08-09.  The occupation of Dino Beltran is associate professor.  The hobby of Dino Beltran is shogi.
Thought 3: The brother of Dino Beltran is Orlando Beltran, so the answer is Orlando Beltran.
Action 3: Finish[Orlando Beltran].

Example 8:
Question: What is the date of birth of the person whose hobby is meteorology?
Thought 1: First, I need to search for people whose hobby is meteorology.
Action 1: Search[meteorology].
Observation 1: (1) Alison Smock (2) Barabara Beltran
Thought 2: People whose hobby is meteorology are Alison Smock, Barabara Beltran. Now I need to retrieve article about Alison Smock and Barabara Beltran and find their date of birth.
Action 2: RetrieveArticle[Alison Smock].
Observation 2: # Alison Smock ## Family The children of Alison Smock are Shelli Beltran, Stacia Toombs.  The daughters of Alison Smock are Shelli Beltran, Stacia Toombs.  The husband of Alison Smock is Williams Smock.  ## Friends The friends of Alison Smock are Eli Smock, Lannie Smock, Virgil Hackworth.  ## Attributes The date of birth of Alison Smock is 0929-10-28.  The occupation of Alison Smock is health promotion specialist.  The hobby of Alison Smock is meteorology.
Thought 3: The date of birth of Alison Smock is 0929-10-28. Now I need to retrieve article about Barabara Beltran and find her date of birth.
Action 3: RetrieveArticle[Barabara Beltran].
Observation 3: # Barabara Beltran ## Family The parents of Barabara Beltran are Dino Beltran, Shelli Beltran.  Barabara Beltran's siblings are Aida Wang, Vicki Hackworth.  The sisters of Barabara Beltran are Aida Wang, Vicki Hackworth.  The mother of Barabara Beltran is Shelli Beltran.  The father of Barabara Beltran is Dino Beltran.  ## Friends ## Attributes The date of birth of Barabara Beltran is 0989-06-11.  The occupation of Barabara Beltran is broadcast engineer.  The hobby of Barabara Beltran is meteorology.
Thought 4: The date of birth of Barabara Beltran is 0989-06-11. So the answer is 0929-10-28, 0989-06-11.
Action 4: Finish[0929-10-28{constants.answer_sep}0989-06-11].

Example 9:
Question: Who is the cousin of the person whose occupation is broadcast engineer?
Thought 1: First, I need to search for people whose occupation is broadcast engineer.
Action 1: Search[broadcast engineer].
Observation 1: # Barabara Beltran ## Family The parents of Barabara Beltran are Dino Beltran, Shelli Beltran.  Barabara Beltran's siblings are Aida Wang, Vicki Hackworth.  The sisters of Barabara Beltran are Aida Wang, Vicki Hackworth.  The mother of Barabara Beltran is Shelli Beltran.  The father of Barabara Beltran is Dino Beltran.  ## Friends ## Attributes The date of birth of Barabara Beltran is 0989-06-11.  The occupation of Barabara Beltran is broadcast engineer.  The hobby of Barabara Beltran is meteorology.
Thought 2: Person whose hobby is broadcast engineer is Barabara Beltran. The cousin of Barabara Beltran is the child of sibling of her parent. Barabara Beltran's parents are Dino Beltran and Shelly Beltran. Now I need to find the sibling of Dino Beltran and Shelly Beltran.
Action 2: RetrieveArticle[Dino Beltran].
Observation 2: # Dino Beltran ## Family The parents of Dino Beltran are Brian Beltran, Daisy Beltran.  Dino Beltran's sibling is Orlando Beltran.  The brother of Dino Beltran is Orlando Beltran.  The mother of Dino Beltran is Daisy Beltran.  The father of Dino Beltran is Brian Beltran.  The children of Dino Beltran are Aida Wang, Barabara Beltran, Vicki Hackworth.  The daughters of Dino Beltran are Aida Wang, Barabara Beltran, Vicki Hackworth.  The wife of Dino Beltran is Shelli Beltran.  ## Friends The friend of Dino Beltran is Alvaro Smock.  ## Attributes The date of birth of Dino Beltran is 0958-08-09.  The occupation of Dino Beltran is associate professor.  The hobby of Dino Beltran is shogi.
Thought 3: The sibling of Dino Beltran is Orlando Beltran. Now I need to retrieve article about Orlando Beltran and find his child.
Action 3: RetrieveArticle[Orlando Beltran].
Observation 3: # Orlando Beltran ## Family The parents of Orlando Beltran are Brian Beltran, Daisy Beltran.  Orlando Beltran's sibling is Dino Beltran.  The brother of Orlando Beltran is Dino Beltran.  The mother of Orlando Beltran is Daisy Beltran.  The father of Orlando Beltran is Brian Beltran.  ## Friends The friends of Orlando Beltran are Virgil Hackworth, Dominique Smock.  ## Attributes The date of birth of Orlando Beltran is 0953-10-23.  The occupation of Orlando Beltran is petroleum engineer.  The hobby of Orlando Beltran is learning.
Thought 4: Orlando Beltran has no child, so no cousin from the side of Dino Beltran. Now I need to retrieve article about Shelli Beltran and find her sibling.
Action 4: RetrieveArticle[Shelli Beltran].
Observation 4: # Shelli Beltran ## Family The parents of Shelli Beltran are Alison Smock, Williams Smock.  Shelli Beltran's sibling is Stacia Toombs.  The sister of Shelli Beltran is Stacia Toombs.  The mother of Shelli Beltran is Alison Smock.  The father of Shelli Beltran is Williams Smock.  The children of Shelli Beltran are Aida Wang, Barabara Beltran, Vicki Hackworth.  The daughters of Shelli Beltran are Aida Wang, Barabara Beltran, Vicki Hackworth.  The husband of Shelli Beltran is Dino Beltran.  ## Friends The friends of Shelli Beltran are Brian Beltran, Eli Smock, Isiah Lutz, Leslee Toombs, Lesley Lutz, Ryan Wang.  ## Attributes The date of birth of Shelli Beltran is 0958-03-08.  The occupation of Shelli Beltran is occupational therapist.  The hobby of Shelli Beltran is sociology.
Thought 5: The sibling of Shelli Beltran is Stacia Toombs. Now I need to retrieve article about Stacia Toombs and find her child, to get the cousin of Barabara Beltran.
Action 5: RetrieveArticle[Stacia Toombs].
Observation 5: # Stacia Toombs ## Family The parents of Stacia Toombs are Alison Smock, Williams Smock.  Stacia Toombs's sibling is Shelli Beltran.  The sister of Stacia Toombs is Shelli Beltran.  The mother of Stacia Toombs is Alison Smock.  The father of Stacia Toombs is Williams Smock.  The child of Stacia Toombs is Leslee Toombs.  The daughter of Stacia Toombs is Leslee Toombs.  The husband of Stacia Toombs is Wilbert Toombs.  ## Friends The friends of Stacia Toombs are Brian Beltran, Isiah Lutz, Leeann Hackworth, Lesley Lutz, Ryan Wang.  ## Attributes The date of birth of Stacia Toombs is 0959-03-22.  The occupation of Stacia Toombs is actuary.  The hobby of Stacia Toombs is finance.
Thought 6: The child of Stacia Toombs is Leslee Toombs. So the cousin of Barabara Beltran is Leslee Toombs. The answer is Leslee Toombs.
Action 6: Finish[Leslee Toombs].

Example 10:
Question: Who is the great-granddaughter of the person whose hobby is biology?
Thought 1: First, I need to search for people whose hobby is biology.
Action 1: Search[biology].
Observation 1: (1) Alvaro Smock
Thought 2: Person whose hobby is biology is Alvaro Smock. Now I need to retrieve article about Alvaro Smock and find his great-granddaughter. A great-granddaughter is the daughter of the child of the child of the person.
Action 2: RetrieveArticle[Alvaro Smock].
Observation 2: # Alvaro Smock ## Family The children of Alvaro Smock are Eli Smock, Gene Smock.  The sons of Alvaro Smock are Eli Smock, Gene Smock.  The wife of Alvaro Smock is Lannie Smock.  ## Friends The friends of Alvaro Smock are Dino Beltran, Gene Smock, Aida Wang.  ## Attributes The date of birth of Alvaro Smock is 0867-07-12.  The occupation of Alvaro Smock is osteopath.  The hobby of Alvaro Smock is biology.
Thought 3: The children of Alvaro Smock are Eli Smock, Gene Smock. Now I need to retrieve article about Eli Smock and Gene Smock and find their child.
Action 3: RetrieveArticle[Eli Smock].
Observation 3: # Eli Smock ## Family The parents of Eli Smock are Alvaro Smock, Lannie Smock.  Eli Smock's sibling is Gene Smock.  The brother of Eli Smock is Gene Smock.  The mother of Eli Smock is Lannie Smock.  The father of Eli Smock is Alvaro Smock.  ## Friends The friends of Eli Smock are Leisa Lutz, Shelli Beltran, Vicki Hackworth, Virgil Hackworth, Alison Smock, Brian Beltran.  ## Attributes The date of birth of Eli Smock is 0901-01-18.  The occupation of Eli Smock is retail manager.  The hobby of Eli Smock is tether car.
Thought 4: Eli Smock has no child. Now I need to retrieve article about Gene Smock and find his child.
Action 4: RetrieveArticle[Gene Smock].
Observation 4: # Gene Smock ## Family The parents of Gene Smock are Alvaro Smock, Lannie Smock.  Gene Smock's sibling is Eli Smock.  The brother of Gene Smock is Eli Smock.  The mother of Gene Smock is Lannie Smock.  The father of Gene Smock is Alvaro Smock.  The child of Gene Smock is Williams Smock.  The son of Gene Smock is Williams Smock.  The wife of Gene Smock is Dominique Smock.  ## Friends The friends of Gene Smock are Leeann Hackworth, Leisa Lutz, Ricardo Hackworth, Alvaro Smock, Dominique Smock.  ## Attributes The date of birth of Gene Smock is 0898-08-16.  The occupation of Gene Smock is immunologist.  The hobby of Gene Smock is architecture.
Thought 5: The child of Gene Smock is Williams Smock. Now I need to retrieve article about Williams Smock and find his daughter, to get the great-granddaughter of Alvaro Smock.
Action 5: RetrieveArticle[Williams Smock].
Observation 5: # Williams Smock ## Family The parents of Williams Smock are Dominique Smock, Gene Smock.  The mother of Williams Smock is Dominique Smock.  The father of Williams Smock is Gene Smock.  The children of Williams Smock are Shelli Beltran, Stacia Toombs.  The daughters of Williams Smock are Shelli Beltran, Stacia Toombs.  The wife of Williams Smock is Alison Smock.  ## Friends The friend of Williams Smock is Lannie Smock.  ## Attributes The date of birth of Williams Smock is 0926-04-04.  The occupation of Williams Smock is clinical biochemist.  The hobby of Williams Smock is social studies.
Thought 6: The daughters of Williams Smock are Shelli Beltran, Stacia Toombs. So the great-granddaughters of Alvaro Smock, whose hobby is biology, are Shelli Beltran, Stacia Toombs. The answer is Shelli Beltran, Stacia Toombs.
Action 6: Finish[Shelli Beltran{constants.answer_sep}Stacia Toombs].
(END OF EXAMPLES)

Now answer the following question:
Question: {{question}}
{{scratchpad}}

\end{lstlisting}

%% file: appendices/additional_results.tex
\section{Additional Results}

\begin{figure*}[t!]
    \centering
    \includegraphics[width=0.9\linewidth]{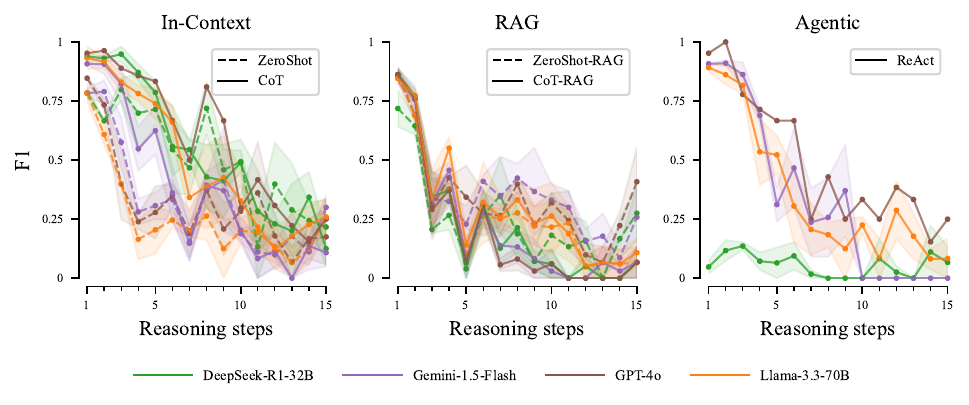}
    \caption{%
    \textbf{
    F1 scores versus question difficulty, measured by \textit{reasoning steps}, for questions with exactly 1 solution.} We observe similar trends as in \cref{fig:exp__f1_v_difficulty}, demonstrating that the number of solutions is not solely responsible for the drop in performance.
    }
    \label{fig:exp__f1_sol1_v_difficulty}
\end{figure*}

\subsection{Multi-Hop RAG Baselines}
\label{app:multi_hop_rag}

\begin{figure*}[b!]
    \centering
    \includegraphics{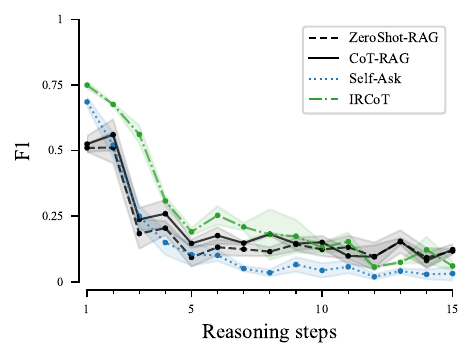}
    \caption{%
    \textbf{F1 scores versus question difficulty, measured by reasoning steps.} \zeroshotrag and \cotrag use the \retriever retriever model (top-k$=4$); \textsc{Self-Ask} and \textsc{IRCoT} use the BM25 retriever (top-k$=5$). For all methods, we use \llama as the generator model. We report mean $\pm$ 1 standard error across 3 dataset generations seeds with universe size $n=50$.
    }
    \label{fig:additional_rag_baselines}
\end{figure*}

We include two RAG baselines that interleave reasoning with retrieval: Self-Ask \citep{press2023measuring} and IRCoT \citep{trivedi2023interleaving}. We use the implementation from FlashRAG (\url{https://github.com/RUC-NLPIR/FlashRAG}) and write few-shot examples suited to \ours. 
\cref{tab:additional_rag_baselines} includes the results of these baselines on the same \ours instances as in \cref{tab:exp__results}. Notably, IRCoT improves over \zeroshotrag and \cotrag, but struggles on questions requiring many reasoning steps, as revealed by \cref{fig:additional_rag_baselines}.

\begin{table}[t!]
    \centering
    \caption{\textbf{F1 scores (in \%) for \llama with RAG prompting.} \zeroshotrag and \cotrag use the \retriever retriever model (top-k$=4$); \textsc{Self-Ask} and \textsc{IRCoT} use the BM25 retriever (top-k$=5$). We report mean $\pm$ 1 standard error across 3 dataset generation seeds.}
    \begin{tabular}{ccccc}
        \toprule
        \textbf{Universe Size $n$} & \textbf{\zeroshotrag} & \textbf{\cotrag} & \textbf{\textsc{Self-Ask}} & \textbf{\textsc{IRCoT}} \\
        \midrule
        50 & 17.55 ± 2.20 & 20.01 ± 1.81 & 15.77 ± 3.22 & 23.93 ± 0.93
        \\
        \bottomrule
    \end{tabular}
    \label{tab:additional_rag_baselines}
\end{table}

\subsection{Article Rephrasing}
\label{app:article-rephrasing}
To explore how LLMs can be used to improve the realism of our templated articles, we instruct \llama to rephrase articles in corpora of size $n=50$ from \cref{fig:exp__f1_v_difficulty}. Our first, ``short'' prompt instructs the LLM to condense the templated articles, while still retaining all factual information:

\begin{lstlisting}[basicstyle=\ttfamily,breaklines=true]
Shuffle and rephrase the following wikipedia-like article. Keep ALL facts 
like name, relation, date, occupation, hobby and gender exactly as stated. Do
not add ANY new information.
\end{lstlisting}

Our second, ``long'' prompt permits the LLM to introduce new facts, without contradicting existing facts:

\begin{lstlisting}[basicstyle=\ttfamily,breaklines=true]
Transform the following factual article into an engaging narrative profile:

1. Add colorful descriptions and personality traits that might be inferred
2. Create a vivid backstory about how relationships formed
3. Elaborate on the person's career path and achievements
4. Describe their hobbies in rich detail, including when they might have started them
5. Include hypothetical quotes from family members
6. Imagine and describe the person's daily routine
7. Add details about where they might live and their home environment
8. Write in a warm, personal tone as if you've known the subject for years
9. Maintain a natural flow

When transforming the article, ensure the following:
1. All factual information remains 100%
2. No new information is added or implied 
3. The content is presented in a natural, flowing narrative rather than just shuffling bullet points
4. Different sections are reorganized in a way that still makes logical sense
5. Sentence structures and vocabulary are varied from the original

This is the article to transform:
\end{lstlisting}

\begin{figure}[t!]
    \centering
    \begin{tabular}{cc}
        \includegraphics[width=0.45\textwidth]{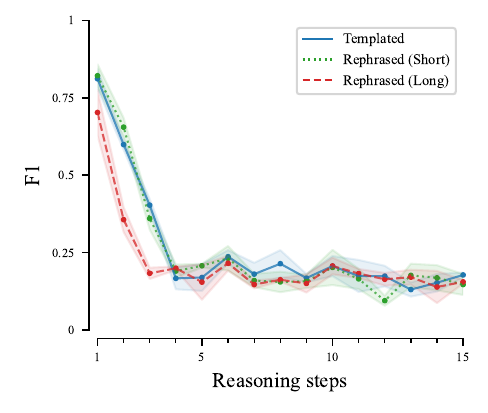} &  
        \includegraphics[width=0.45\textwidth]{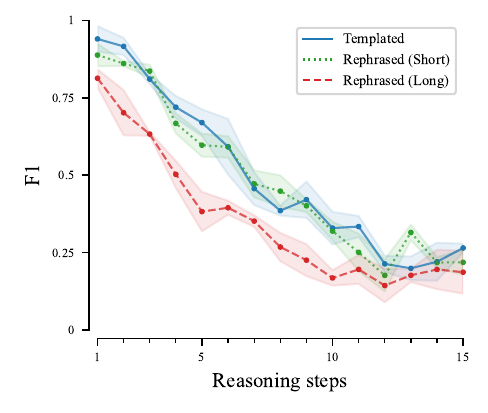} \\
        (a) \zeroshot Prompting &
        (b) \CoT Prompting
    \end{tabular}
    \caption{\tbf{F1 scores versus question difficulty, measured by reasoning steps, for various article styles}.
    We rephrase templated articles generated by \ours with \llama.
    Then for evaluation, we replace templated articles with LLM-rephrased articles in \zeroshot and \CoT prompts from \cref{sub:zeroshot_simple} and \cref{sub:cot_simple} respectively.
    As in \cref{fig:exp__f1_v_difficulty}, we report mean $\pm$ 1 standard error across 3 dataset generation seeds for universe size $n=50$.     
    We use evaluation sampling hyper-parameters from \cref{sec:main_results}.
    }
    \label{fig:llm_rephrasing}
\end{figure}

To generate articles of either styles, we prompt \llama with temperature $=0.7$, top-p $=0.9$, and max output tokens $=1024$.
We include an example short article in \cref{sub:example_templated} and an example long article in \cref{sub:example_rephrased_long}. \cref{fig:llm_rephrasing} shows plots of F1 scores vs question difficulty of \llama using \zeroshot and \CoT prompting on these LLM-rephrased articles.
We find that regardless of the article style---templated or LLM-rephrased---F1 scores decline as the number of reasoning steps increases.

\subsubsection{Example of Templated Article} \label{sub:example_templated}

\begin{lstlisting}[basicstyle=\ttfamily,breaklines=true]
# Alison Smock

## Family
The daughters of Alison Smock are Cythia Smock, Shelli Beltran, Stacia Toombs.
The husband of Alison Smock is Williams Smock.

## Friends
The friends of Alison Smock are Cortney Parmer, Jamison Baptiste, Wilbert Toombs.

## Attributes
The date of birth of Alison Smock is 0929-10-28.
The occupation of Alison Smock is broadcast engineer.
The hobby of Alison Smock is meteorology.
The gender of Alison Smock is female.
\end{lstlisting}

\subsubsection{Example of Short Rephrased Article} \label{sub:example_rephrased_short}

The following article was generated with \llama using 118 output tokens.

\begin{quoting}
Alison Smock, a female born on 0929-10-28, is a broadcast engineer with a hobby of meteorology. She is married to Williams Smock and they have daughters named Cythia Smock, Shelli Beltran, and Stacia Toombs. Alison Smock's social circle includes friends such as Cortney Parmer, Jamison Baptiste, and Wilbert Toombs.
\end{quoting}

\subsubsection{Example of Long Rephrased Article} \label{sub:example_rephrased_long}

The following article was generated with \llama using 669 output tokens. \llama incorrectly stated the date-of-birth (red) and omitted the last names of the friends (orange).

\begin{quoting}
Alison Smock, a brilliant and charismatic broadcast engineer, has been fascinated by the intricacies of technology and the wonders of the atmosphere for as long as anyone can remember. \red{Born on October 28, 1929}, Alison grew up with an insatiable curiosity and a passion for learning that would serve her well throughout her life. As a young girl, she was always tinkering with gadgets and machinery, trying to understand how they worked, and this innate curiosity eventually led her to pursue a career in broadcast engineering.

As she navigated her professional path, Alison met her husband, Williams Smock, and together they built a loving family. \orange{They had three beautiful daughters, Cythia, Shelli, and Stacia}, who brought immense joy and fulfillment to their lives. According to her daughter Cythia, ``Mom was always there for us, supporting us and encouraging us to pursue our dreams, no matter how big or small they seemed." Alison's family was her rock, and she cherished every moment she spent with them.

Alison's career as a broadcast engineer was marked by numerous achievements and accomplishments. She was known for her exceptional problem-solving skills and her ability to stay calm under pressure. Her colleagues admired her dedication and work ethic, and she quickly became an invaluable member of her team. When she wasn't working, Alison loved to indulge in her hobby of meteorology. She found the study of weather patterns and atmospheric conditions absolutely captivating, and she spent hours poring over books and watching documentaries on the subject. As her friend Cortney Parmer recalls, ``Alison would often call me up and say, 'Did you see that storm system moving in? I've been tracking it all day!' She was like a kid in a candy store when it came to weather."

In her free time, Alison enjoyed spending time with her friends, including Cortney, Jamison Baptiste, and Wilbert Toombs. They would often meet up for coffee or go on hikes together, discussing everything from the latest news to their personal lives. Wilbert, who is also the husband of Alison's daughter Stacia, says, ``Alison is like a mother to me. She's always been so kind and supportive, and I admire her intelligence and passion for life."

A typical day for Alison begins early, with a quiet moment of reflection and a cup of coffee on her back porch, surrounded by the beauty of nature. She lives in a cozy home filled with books, gadgets, and mementos from her travels, and she loves to spend time in her garden, tending to her plants and watching the birds. As she gets ready for work, Alison feels grateful for the life she's built and the people who bring joy and love into her world. As her husband Williams says, ``Alison is an amazing woman, and I feel so lucky to have her by my side. She's intelligent, kind, and beautiful, and I admire her every day." With her quick wit, sharp mind, and generous spirit, Alison Smock is a true treasure, and those who know her feel fortunate to have her in their lives.
\end{quoting}

\subsection{Fine-tuning on \ours -- Experiment Details}
\label{sub:finetuning_details}

\ours creates datasets with relationships, names, attributes populated randomly on-demand, ensuring that each dataset instance is unique and robust to LLMs memorizing dataset facts.
Even so, it is possible that fine-tuning on \ours-generated datasets could improve performance by leveraging language and template structure in \ours articles and questions.
Accordingly, we fine-tune LLMs on \ours datasets to  (1) assess the viability of \ours for training LLMs and (2) test the robustness of \ours on-demand datasets from memorization.

We generate 10 new PhantomWiki dataset instances (question depth 20 and universe size 50) amounting to 5K training question-answer pairs.
We then perform full fine-tuning of Qwen2.5-0.5B-Instruct~\citep{yang2024qwen2} and parameter-efficient fine-tuning of Qwen2.5-3B-Instruct with LoRA~\citep{hu2022lora} applied to all linear layers.
We use a node of 4 A100 GPUs each with 80GB GPU VRAM  for each experiment.
For each base model, we employ two popular fine-tuning algorithms:
\begin{enumerate}
    \item The first is \textbf{Group Relative Policy Optimization (GRPO)}~\citep{shao2024deepseekmath}, where we elicit chain-of-thought response from the LLM using our \CoT template (\cref{sub:cot_simple}) and return reward as the F1-score of the generated answer list against the ground-truth answer list.

    We use the \href{https://huggingface.co/docs/trl/main/en/grpo_trainer}{GRPO implementation} from Huggingface's TRL library.
    We set max prompt length $=4096$, sufficient to include our prompt of the 50 articles of the universe, and limit the max completion length to $=128$.
    We reserve 1 GPU for generation with vLLM, and use the other 3 GPUs for training to get $6$ total generations per step of GRPO update.
    For full fine-tuning, we set per-device training batch size $=8$ and gradient accumulation steps $=8$, and for LoRA fine-tuning we set $2$ and $4$ respectively.
    We fine-tune for 3 epochs using the AdamW optimizer with initial learning rate set to $5 \times 10^{-6}$ for full fine-tuning and $10^{-4}$ for LoRA fine-tuning.

    \item The second is \textbf{Supervised Fine-tuning (SFT)}, where we provide all articles and the question using our \zeroshot prompt in \cref{sub:zeroshot_simple} and train on the ground-truth answer list.

    Again we use the \href{https://huggingface.co/docs/trl/main/en/sft_trainer}{SFT implementation} from Huggingface's TRL library and set the max prompt length to $4096$.
    For both full fine-tuning and LoRA fine-tuning, we set per-device training batch size and gradient accumulation steps to $1$.
    We fine-tune for 3 epochs using the AdamW optimizer with initial learning rate set to $2 \times 10^{-5}$ for full fine-tuning and $10^{-4}$ for LoRA fine-tuning.
\end{enumerate}

\begin{figure}[h]
    \centering
    \begin{tabular}{cc}
        \includegraphics[width=0.45\textwidth]{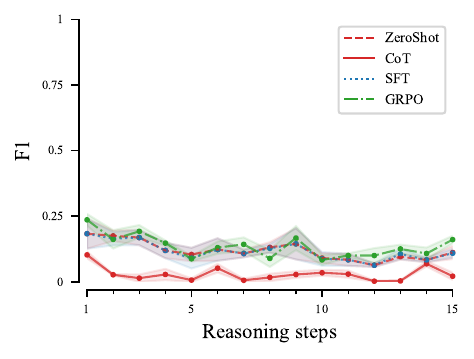} &  
        \includegraphics[width=0.45\textwidth]{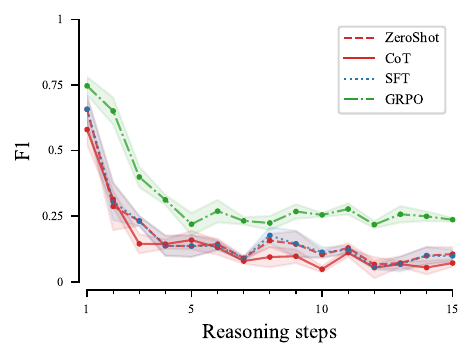} \\
        (a) Qwen2.5-0.5B-Instruct (full fine-tuning) &
        (b) Qwen2.5-3B-Instruct (LoRA fine-tuning)
    \end{tabular}
    \caption{\tbf{F1 scores versus question difficulty, measured by reasoning steps.} We report mean $\pm$ 1 standard error across 3 dataset generation seeds for universe size $n=50$. \zeroshot and \CoT are prompting methods; SFT and GRPO are fine-tuning methods.
    }
    \label{fig:fine_tuning}
\end{figure}

We evaluate these models on the three \ours dataset instances of size $n=50$ and maximum recursion depth 20 (500 questions per dataset instance), the same used in 
\cref{tab:exp__results,fig:exp__f1_v_difficulty}.
\cref{fig:fine_tuning} shows F1 scores as a function of reasoning steps, comparing base LLMs with fine-tuned ones, and \cref{tab:fine_tuning} shows the aggregate F1 scores.
To adhere to the fine-tuning experiment setup, SFT-trained LLMs are evaluated with \zeroshot prompt, and GRPO-trained with \CoT prompt.

For both full fine-tuned Qwen2.5-0.5B-Instruct and LoRA fine-tuned Qwen2.5-3B-Instruct, we see that GRPO improves performance for all question difficulties compared to \CoT and \zeroshot prompting.
We find that SFT does not improve performance over \zeroshot prompting.
Notably, performance of fine-tuned LLMs is still far from optimal and the F1 scores struggle as question difficulty increases.

%% file: appendices/small_corpus.tex
\section{Example of Small Corpus}
\label{app:example-corpus}

We generate a universe of size $n=4$, setting the number of family trees to be one. We include the articles below.

\subsection{Article 1}
\begin{lstlisting}[basicstyle=\ttfamily,breaklines=true]
# Claud Colin

## Family
The brother of Claud Colin is Mckinley Colin.
The mother of Claud Colin is Ramona Colin.
The father of Claud Colin is Danilo Colin.

## Friends
The friend of Claud Colin is Danilo Colin.

## Attributes
The date of birth of Claud Colin is 0241-12-06.
The occupation of Claud Colin is academic librarian.
The hobby of Claud Colin is amateur astronomy.
The gender of Claud Colin is male.
\end{lstlisting}

\subsection{Article 2}
\begin{lstlisting}[basicstyle=\ttfamily,breaklines=true]
# Danilo Colin

## Family
The sons of Danilo Colin are Claud Colin, Mckinley Colin.
The wife of Danilo Colin is Ramona Colin.

## Friends
The friends of Danilo Colin are Mckinley Colin, Ramona Colin, Claud Colin.

## Attributes
The date of birth of Danilo Colin is 0219-08-09.
The occupation of Danilo Colin is clinical research associate.
The hobby of Danilo Colin is crystals.
The gender of Danilo Colin is male.
\end{lstlisting}

\newpage
\subsection{Article 3}
\begin{lstlisting}[basicstyle=\ttfamily,breaklines=true]
# Mckinley Colin

## Family
The brother of Mckinley Colin is Claud Colin.
The mother of Mckinley Colin is Ramona Colin.
The father of Mckinley Colin is Danilo Colin.

## Friends
The friends of Mckinley Colin are Ramona Colin, Danilo Colin.

## Attributes
The date of birth of Mckinley Colin is 0246-10-18.
The occupation of Mckinley Colin is museum curator.
The hobby of Mckinley Colin is stamp collecting.
The gender of Mckinley Colin is male.
\end{lstlisting}

\subsection{Article 4}
\begin{lstlisting}[basicstyle=\ttfamily,breaklines=true]
# Ramona Colin

## Family
The sons of Ramona Colin are Claud Colin, Mckinley Colin.
The husband of Ramona Colin is Danilo Colin.

## Friends
The friends of Ramona Colin are Danilo Colin, Mckinley Colin.

## Attributes
The date of birth of Ramona Colin is 0219-09-08.
The occupation of Ramona Colin is technical sales engineer.
The hobby of Ramona Colin is trainspotting.
The gender of Ramona Colin is female.

\end{lstlisting}

%% file: main.bbl
\begin{thebibliography}{75}
\providecommand{\natexlab}[1]{#1}
\providecommand{\url}[1]{\texttt{#1}}
\expandafter\ifx\csname urlstyle\endcsname\relax
  \providecommand{\doi}[1]{doi: #1}\else
  \providecommand{\doi}{doi: \begingroup \urlstyle{rm}\Url}\fi

\bibitem[Agarwal et~al.(2020)Agarwal, Ge, Shakeri, and Al-Rfou]{agarwal2020knowledge}
Agarwal, O., Ge, H., Shakeri, S., and Al-Rfou, R.
\newblock Knowledge graph based synthetic corpus generation for knowledge-enhanced language model pre-training.
\newblock \emph{arXiv preprint arXiv:2010.12688}, 2020.

\bibitem[{AlphaProof \& AlphaGeometry}(2024)]{alphaproof2024}
{AlphaProof \& AlphaGeometry}.
\newblock {AI} achieves silver-medal standard solving {International Mathematical Olympiad} problems.
\newblock \href{https://deepmind.google/discover/blog/ai-solves-imo-problems-at-silver-medal-level/}{https://deepmind.google/discover/blog/ai-solves-imo-problems-at-silver-medal-level/}, 2024.
\newblock Accessed: 2025-01-25.

\bibitem[An et~al.(2023)An, Gong, Zhong, Zhao, Li, Zhang, Kong, and Qiu]{an2023eval}
An, C., Gong, S., Zhong, M., Zhao, X., Li, M., Zhang, J., Kong, L., and Qiu, X.
\newblock {L-Eval}: Instituting standardized evaluation for long context language models.
\newblock \emph{arXiv preprint arXiv:2307.11088}, 2023.

\bibitem[Bai et~al.(2023)Bai, Lv, Zhang, Lyu, Tang, Huang, Du, Liu, Zeng, Hou, et~al.]{bai2023longbench}
Bai, Y., Lv, X., Zhang, J., Lyu, H., Tang, J., Huang, Z., Du, Z., Liu, X., Zeng, A., Hou, L., et~al.
\newblock {LongBench}: A bilingual, multitask benchmark for long context understanding.
\newblock \emph{arXiv preprint arXiv:2308.14508}, 2023.

\bibitem[Borgeaud et~al.(2022)Borgeaud, Mensch, Hoffmann, Cai, Rutherford, Millican, Van Den~Driessche, Lespiau, Damoc, Clark, et~al.]{borgeaud2022improving}
Borgeaud, S., Mensch, A., Hoffmann, J., Cai, T., Rutherford, E., Millican, K., Van Den~Driessche, G.~B., Lespiau, J.-B., Damoc, B., Clark, A., et~al.
\newblock Improving language models by retrieving from trillions of tokens.
\newblock In \emph{International conference on machine learning}, pp.\  2206--2240. PMLR, 2022.

\bibitem[Chen et~al.(2020)Chen, Zha, Chen, Xiong, Wang, and Wang]{chen2020hybridqa}
Chen, W., Zha, H., Chen, Z., Xiong, W., Wang, H., and Wang, W.
\newblock {HybridQA}: A dataset of multi-hop question answering over tabular and textual data.
\newblock \emph{arXiv preprint arXiv:2004.07347}, 2020.

\bibitem[Cobbe et~al.(2021)Cobbe, Kosaraju, Bavarian, Chen, Jun, Kaiser, Plappert, Tworek, Hilton, Nakano, Hesse, and Schulman]{cobbe2021training}
Cobbe, K., Kosaraju, V., Bavarian, M., Chen, M., Jun, H., Kaiser, L., Plappert, M., Tworek, J., Hilton, J., Nakano, R., Hesse, C., and Schulman, J.
\newblock Training verifiers to solve math word problems.
\newblock \emph{arXiv preprint arXiv:2110.14168}, 2021.

\bibitem[Cohen et~al.(2024)Cohen, Biran, Yoran, Globerson, and Geva]{cohen2024evaluating}
Cohen, R., Biran, E., Yoran, O., Globerson, A., and Geva, M.
\newblock Evaluating the ripple effects of knowledge editing in language models.
\newblock \emph{Transactions of the Association for Computational Linguistics}, 12:\penalty0 283--298, 2024.

\bibitem[{\relax DeepSeek AI}(2024)]{shao2024deepseekmath}
{\relax DeepSeek AI}.
\newblock {DeepSeekMath}: Pushing the limits of mathematical reasoning in open language models.
\newblock \emph{arXiv preprint arXiv:2402.03300}, 2024.

\bibitem[{\relax DeepSeek AI}(2025)]{guo2025deepseek}
{\relax DeepSeek AI}.
\newblock {DeepSeek-R1}: Incentivizing reasoning capability in {LLMs} via reinforcement learning.
\newblock \emph{arXiv preprint arXiv:2501.12948}, 2025.

\bibitem[Dua et~al.(2019)Dua, Wang, Dasigi, Stanovsky, Singh, and Gardner]{dua2019drop}
Dua, D., Wang, Y., Dasigi, P., Stanovsky, G., Singh, S., and Gardner, M.
\newblock {DROP}: A reading comprehension benchmark requiring discrete reasoning over paragraphs.
\newblock \emph{arXiv preprint arXiv:1903.00161}, 2019.

\bibitem[Elazar et~al.(2021)Elazar, Kassner, Ravfogel, Ravichander, Hovy, Sch{\"u}tze, and Goldberg]{elazar2021measuring}
Elazar, Y., Kassner, N., Ravfogel, S., Ravichander, A., Hovy, E., Sch{\"u}tze, H., and Goldberg, Y.
\newblock Measuring and improving consistency in pretrained language models.
\newblock \emph{Transactions of the Association for Computational Linguistics}, 9:\penalty0 1012--1031, 2021.

\bibitem[Feldman \& El-Yaniv(2019)Feldman and El-Yaniv]{feldman2019multi}
Feldman, Y. and El-Yaniv, R.
\newblock Multi-hop paragraph retrieval for open-domain question answering.
\newblock \emph{arXiv preprint arXiv:1906.06606}, 2019.

\bibitem[Flajolet \& Sedgewick(2009)Flajolet and Sedgewick]{flajolet2009analytic}
Flajolet, P. and Sedgewick, R.
\newblock \emph{Analytic Combinatorics}.
\newblock Cambridge University Press, 2009.
\newblock ISBN 9781139477161.
\newblock URL \url{https://books.google.com/books?id=0h-4QcA1c1QC}.

\bibitem[{Gemini Team, Google}(2024)]{google2024gemini}
{Gemini Team, Google}.
\newblock {Gemini~1.5}: Unlocking multimodal understanding across millions of tokens of context.
\newblock \emph{arXiv preprint arXiv:2403.05530}, 2024.

\bibitem[Guu et~al.(2020)Guu, Lee, Tung, Pasupat, and Chang]{guu2020retrieval}
Guu, K., Lee, K., Tung, Z., Pasupat, P., and Chang, M.
\newblock Retrieval augmented language model pre-training.
\newblock In \emph{International conference on machine learning}, pp.\  3929--3938. PMLR, 2020.

\bibitem[Han et~al.(2022)Han, Schoelkopf, Zhao, Qi, Riddell, Zhou, Coady, Peng, Qiao, Benson, Sun, Wardle-Solano, Szabo, Zubova, Burtell, Fan, Liu, Wong, Sailor, Ni, Nan, Kasai, Yu, Zhang, Fabbri, Kryscinski, Yavuz, Liu, Lin, Joty, Zhou, Xiong, Ying, Cohan, and Radev]{han2022folio}
Han, S., Schoelkopf, H., Zhao, Y., Qi, Z., Riddell, M., Zhou, W., Coady, J., Peng, D., Qiao, Y., Benson, L., Sun, L., Wardle-Solano, A., Szabo, H., Zubova, E., Burtell, M., Fan, J., Liu, Y., Wong, B., Sailor, M., Ni, A., Nan, L., Kasai, J., Yu, T., Zhang, R., Fabbri, A.~R., Kryscinski, W., Yavuz, S., Liu, Y., Lin, X.~V., Joty, S., Zhou, Y., Xiong, C., Ying, R., Cohan, A., and Radev, D.
\newblock {FOLIO}: Natural language reasoning with first-order logic.
\newblock \emph{arXiv preprint arXiv:2209.00840}, 2022.

\bibitem[Hendrycks et~al.(2020)Hendrycks, Burns, Basart, Zou, Mazeika, Song, and Steinhardt]{hendrycks2020measuring}
Hendrycks, D., Burns, C., Basart, S., Zou, A., Mazeika, M., Song, D., and Steinhardt, J.
\newblock Measuring massive multitask language understanding.
\newblock \emph{arXiv preprint arXiv:2009.03300}, 2020.

\bibitem[Ho et~al.(2020)Ho, Nguyen, Sugawara, and Aizawa]{ho2020constructing}
Ho, X., Nguyen, A.-K.~D., Sugawara, S., and Aizawa, A.
\newblock Constructing a multi-hop {QA} dataset for comprehensive evaluation of reasoning steps.
\newblock \emph{arXiv preprint arXiv:2011.01060}, 2020.

\bibitem[Hohenecker \& Lukasiewicz(2020)Hohenecker and Lukasiewicz]{hohenecker2020ontology}
Hohenecker, P. and Lukasiewicz, T.
\newblock Ontology reasoning with deep neural networks.
\newblock \emph{Journal of Artificial Intelligence Research}, 68:\penalty0 503--540, 2020.

\bibitem[Hopcroft et~al.(2001)Hopcroft, Motwani, and Ullman]{hopcroft2001introduction}
Hopcroft, J.~E., Motwani, R., and Ullman, J.~D.
\newblock Introduction to automata theory, languages, and computation.
\newblock \emph{Acm Sigact News}, 32\penalty0 (1):\penalty0 60--65, 2001.

\bibitem[Hsia et~al.(2024)Hsia, Shaikh, Wang, and Neubig]{hsia2024ragged}
Hsia, J., Shaikh, A., Wang, Z., and Neubig, G.
\newblock {RAGGED}: Towards informed design of retrieval augmented generation systems.
\newblock \emph{arXiv preprint arXiv:2403.09040}, 2024.

\bibitem[Hsieh et~al.(2024)Hsieh, Sun, Kriman, Acharya, Rekesh, Jia, Zhang, and Ginsburg]{hsieh2024ruler}
Hsieh, C.-P., Sun, S., Kriman, S., Acharya, S., Rekesh, D., Jia, F., Zhang, Y., and Ginsburg, B.
\newblock {RULER}: What's the real context size of your long-context language models?
\newblock \emph{COLM}, 2024.

\bibitem[Hu et~al.(2022)Hu, Shen, Wallis, Allen-Zhu, Li, Wang, Wang, and Chen]{hu2022lora}
Hu, E.~J., Shen, Y., Wallis, P., Allen-Zhu, Z., Li, Y., Wang, S., Wang, L., and Chen, W.
\newblock {LoRA}: Low-rank adaptation of large language models.
\newblock \emph{ICLR}, 1\penalty0 (2):\penalty0 3, 2022.

\bibitem[Jin et~al.(2024)Jin, Zhu, Yang, Zhang, and Dou]{jin2024flashrag}
Jin, J., Zhu, Y., Yang, X., Zhang, C., and Dou, Z.
\newblock {FlashRAG}: A modular toolkit for efficient retrieval-augmented generation research.
\newblock \emph{arXiv preprint arXiv:2405.13576}, 2024.

\bibitem[Karpukhin et~al.(2020)Karpukhin, O{\u{g}}uz, Min, Lewis, Wu, Edunov, Chen, and Yih]{karpukhin2020dense}
Karpukhin, V., O{\u{g}}uz, B., Min, S., Lewis, P., Wu, L., Edunov, S., Chen, D., and Yih, W.-t.
\newblock Dense passage retrieval for open-domain question answering.
\newblock \emph{arXiv preprint arXiv:2004.04906}, 2020.

\bibitem[Kasai et~al.(2024)Kasai, Sakaguchi, Takahashi, Bras, Asai, Yu, Radev, Smith, Choi, and Inui]{kasai2024realtime}
Kasai, J., Sakaguchi, K., Takahashi, Y., Bras, R.~L., Asai, A., Yu, X., Radev, D., Smith, N.~A., Choi, Y., and Inui, K.
\newblock {REALTIME QA}: what's the answer right now?
\newblock \emph{Advances in Neural Information Processing Systems}, 36, 2024.

\bibitem[Lattimer et~al.(2024)Lattimer, Gangal, McDonald, and Yang]{lattimer2024sparse}
Lattimer, B.~M., Gangal, V., McDonald, R., and Yang, Y.
\newblock Sparse rewards can self-train dialogue agents.
\newblock \emph{arXiv preprint arXiv:2409.04617}, 2024.

\bibitem[Lewis et~al.(2020)Lewis, Perez, Piktus, Petroni, Karpukhin, Goyal, K{\"u}ttler, Lewis, tau Yih, Rockt{\"a}schel, Riedel, and Kiela]{lewis2020retrieval}
Lewis, P., Perez, E., Piktus, A., Petroni, F., Karpukhin, V., Goyal, N., K{\"u}ttler, H., Lewis, M., tau Yih, W., Rockt{\"a}schel, T., Riedel, S., and Kiela, D.
\newblock Retrieval-augmented generation for knowledge-intensive nlp tasks.
\newblock \emph{Advances in Neural Information Processing Systems}, 33:\penalty0 9459--9474, 2020.

\bibitem[Li et~al.(2023)Li, Wang, Zheng, and Zhang]{li2023loogle}
Li, J., Wang, M., Zheng, Z., and Zhang, M.
\newblock {LooGLE}: Can long-context language models understand long contexts?
\newblock \emph{arXiv preprint arXiv:2311.04939}, 2023.

\bibitem[Liu et~al.(2020)Liu, Cui, Liu, Huang, Wang, and Zhang]{liu2020logiqa}
Liu, J., Cui, L., Liu, H., Huang, D., Wang, Y., and Zhang, Y.
\newblock Logiqa: A challenge dataset for machine reading comprehension with logical reasoning.
\newblock \emph{arXiv preprint arXiv:2007.08124}, 2020.

\bibitem[{\relax Llama Team -- AI at Meta}(2024)]{dubey2024llama}
{\relax Llama Team -- AI at Meta}.
\newblock The {Llama~3} herd of models.
\newblock \emph{arXiv preprint arXiv:2407.21783}, 2024.

\bibitem[Mao et~al.(2024)Mao, Luo, Zhang, Luo, Cao, Zhang, Hao, Chen, Jiang, Liu, Wang, Huang, Tan, Jie, Li, Liu, Zhang, and Li]{mao2024xrag}
Mao, Q., Luo, Y., Zhang, Q., Luo, Y., Cao, Z., Zhang, J., Hao, H., Chen, Z., Jiang, W., Liu, J., Wang, X., Huang, Z., Tan, Z., Jie, S., Li, B., Liu, X., Zhang, R., and Li, J.
\newblock {XRAG}: {eXamining} the core--benchmarking foundational components in advanced retrieval-augmented generation.
\newblock \emph{arXiv preprint arXiv:2412.15529}, 2024.

\bibitem[Meng et~al.(2022)Meng, Bau, Andonian, and Belinkov]{meng2022locating}
Meng, K., Bau, D., Andonian, A., and Belinkov, Y.
\newblock Locating and editing factual associations in gpt.
\newblock \emph{Advances in Neural Information Processing Systems}, 35:\penalty0 17359--17372, 2022.

\bibitem[Min et~al.(2019)Min, Chen, Hajishirzi, and Zettlemoyer]{min2019discrete}
Min, S., Chen, D., Hajishirzi, H., and Zettlemoyer, L.
\newblock A discrete hard em approach for weakly supervised question answering.
\newblock \emph{arXiv preprint arXiv:1909.04849}, 2019.

\bibitem[Mirzadeh et~al.(2024)Mirzadeh, Alizadeh, Shahrokhi, Tuzel, Bengio, and Farajtabar]{mirzadeh2024gsm}
Mirzadeh, I., Alizadeh, K., Shahrokhi, H., Tuzel, O., Bengio, S., and Farajtabar, M.
\newblock {GSM-Symbolic}: Understanding the limitations of mathematical reasoning in large language models.
\newblock \emph{arXiv preprint arXiv:2410.05229}, 2024.

\bibitem[Monteiro et~al.(2024)Monteiro, Noel, Marcotte, Mudumba, Zantedeschi, Vazquez, Chapados, Pal, and Taslakian]{monteiro2024repliqa}
Monteiro, J., Noel, P.-A., Marcotte, E., Mudumba, S.~R., Zantedeschi, V., Vazquez, D., Chapados, N., Pal, C., and Taslakian, P.
\newblock {RepLiQA}: A question-answering dataset for benchmarking {LLMs} on unseen reference content.
\newblock \emph{Advances in Neural Information Processing Systems}, 37:\penalty0 24242--24276, 2024.

\bibitem[Muennighoff et~al.(2022)Muennighoff, Tazi, Magne, and Reimers]{muennighoff2022mteb}
Muennighoff, N., Tazi, N., Magne, L., and Reimers, N.
\newblock {MTEB}: Massive text embedding benchmark.
\newblock \emph{arXiv preprint arXiv:2210.07316}, 2022.

\bibitem[{\relax OpenAI}(2024)]{hurst2024gpt}
{\relax OpenAI}.
\newblock Gpt-4o system card.
\newblock \emph{arXiv preprint arXiv:2410.21276}, 2024.

\bibitem[Patil et~al.(2023)Patil, Zhang, Wang, and Gonzalez]{patil2023gorilla}
Patil, S.~G., Zhang, T., Wang, X., and Gonzalez, J.~E.
\newblock {Gorilla}: Large language model connected with massive {APIs}.
\newblock \emph{arXiv preprint arXiv:2305.15334}, 2023.

\bibitem[Petroni et~al.(2020)Petroni, Piktus, Fan, Lewis, Yazdani, Cao, Thorne, Jernite, Karpukhin, Maillard, Plachouras, Rocktäschel, and Riedel]{petroni2020kilt}
Petroni, F., Piktus, A., Fan, A., Lewis, P., Yazdani, M., Cao, N.~D., Thorne, J., Jernite, Y., Karpukhin, V., Maillard, J., Plachouras, V., Rocktäschel, T., and Riedel, S.
\newblock {KILT}: a benchmark for knowledge intensive language tasks.
\newblock \emph{arXiv preprint arXiv:2009.02252}, 2020.

\bibitem[Press et~al.(2023)Press, Zhang, Min, Schmidt, Smith, and Lewis]{press2023measuring}
Press, O., Zhang, M., Min, S., Schmidt, L., Smith, N.~A., and Lewis, M.
\newblock Measuring and narrowing the compositionality gap in language models.
\newblock In \emph{Findings of the Association for Computational Linguistics: EMNLP 2023}, pp.\  5687--5711, 2023.

\bibitem[{\relax Qwen Team}(2024)]{yang2024qwen2}
{\relax Qwen Team}.
\newblock {Qwen 2.5 Technical Report}.
\newblock \emph{arXiv preprint arXiv:2412.15115}, 2024.

\bibitem[Rajpurkar(2016)]{rajpurkar2016squad}
Rajpurkar, P.
\newblock {SQuAD}: 100,000+ questions for machine comprehension of text.
\newblock \emph{arXiv preprint arXiv:1606.05250}, 2016.

\bibitem[Rau et~al.(2024)Rau, D{\'e}jean, Chirkova, Formal, Wang, Nikoulina, and Clinchant]{rau2024bergen}
Rau, D., D{\'e}jean, H., Chirkova, N., Formal, T., Wang, S., Nikoulina, V., and Clinchant, S.
\newblock {BERGEN}: A benchmarking library for retrieval-augmented generation.
\newblock \emph{arXiv preprint arXiv:2407.01102}, 2024.

\bibitem[Saad-Falcon et~al.(2023)Saad-Falcon, Khattab, Potts, and Zaharia]{saad2023ares}
Saad-Falcon, J., Khattab, O., Potts, C., and Zaharia, M.
\newblock {ARES}: An automated evaluation framework for retrieval-augmented generation systems.
\newblock \emph{arXiv preprint arXiv:2311.09476}, 2023.

\bibitem[Sakaguchi et~al.(2021)Sakaguchi, Bras, Bhagavatula, and Choi]{sakaguchi2021winogrande}
Sakaguchi, K., Bras, R.~L., Bhagavatula, C., and Choi, Y.
\newblock {WinoGrande}: An adversarial {Winograd} schema challenge at scale.
\newblock \emph{Communications of the ACM}, 64\penalty0 (9):\penalty0 99--106, 2021.

\bibitem[Saparov \& He(2022)Saparov and He]{saparov2022language}
Saparov, A. and He, H.
\newblock Language models are greedy reasoners: A systematic formal analysis of chain-of-thought.
\newblock \emph{arXiv preprint arXiv:2210.01240}, 2022.

\bibitem[Shao et~al.(2024)Shao, Jiang, Kanell, Xu, Khattab, and Lam]{shao2024assisting}
Shao, Y., Jiang, Y., Kanell, T.~A., Xu, P., Khattab, O., and Lam, M.~S.
\newblock Assisting in writing {Wikipedia}-like articles from scratch with large language models.
\newblock \emph{arXiv preprint arXiv:2402.14207}, 2024.

\bibitem[Shi et~al.(2023)Shi, Min, Yasunaga, Seo, James, Lewis, Zettlemoyer, and Yih]{shi2023replug}
Shi, W., Min, S., Yasunaga, M., Seo, M., James, R., Lewis, M., Zettlemoyer, L., and Yih, W.-t.
\newblock {REPLUG}: Retrieval-augmented black-box language models.
\newblock \emph{arXiv preprint arXiv:2301.12652}, 2023.

\bibitem[Shinn et~al.(2024)Shinn, Cassano, Gopinath, Narasimhan, and Yao]{shinn2024reflexion}
Shinn, N., Cassano, F., Gopinath, A., Narasimhan, K., and Yao, S.
\newblock Reflexion: Language agents with verbal reinforcement learning.
\newblock \emph{Advances in Neural Information Processing Systems}, 36, 2024.

\bibitem[Shridhar et~al.(2020)Shridhar, Yuan, C{\^o}t{\'e}, Bisk, Trischler, and Hausknecht]{shridhar2020alfworld}
Shridhar, M., Yuan, X., C{\^o}t{\'e}, M.-A., Bisk, Y., Trischler, A., and Hausknecht, M.
\newblock Alfworld: Aligning text and embodied environments for interactive learning.
\newblock \emph{arXiv preprint arXiv:2010.03768}, 2020.

\bibitem[Sinha et~al.(2019)Sinha, Sodhani, Dong, Pineau, and Hamilton]{sinha2019clutrr}
Sinha, K., Sodhani, S., Dong, J., Pineau, J., and Hamilton, W.~L.
\newblock {CLUTRR}: A diagnostic benchmark for inductive reasoning from text.
\newblock \emph{arXiv preprint arXiv:1908.06177}, 2019.

\bibitem[Sprague et~al.(2023)Sprague, Ye, Bostrom, Chaudhuri, and Durrett]{sprague2023musr}
Sprague, Z., Ye, X., Bostrom, K., Chaudhuri, S., and Durrett, G.
\newblock {MuSR}: Testing the limits of chain-of-thought with multistep soft reasoning.
\newblock \emph{arXiv preprint arXiv:2310.16049}, 2023.

\bibitem[Sterling \& Shapiro(1994)Sterling and Shapiro]{sterling1994art}
Sterling, L. and Shapiro, E.~Y.
\newblock \emph{The art of {Prolog}: advanced programming techniques}.
\newblock {MIT} press, 1994.

\bibitem[Su et~al.(2024)Su, Yen, Xia, Shi, Muennighoff, yu~Wang, Liu, Shi, Siegel, Tang, Sun, Yoon, Arik, Chen, and Yu]{su2024bright}
Su, H., Yen, H., Xia, M., Shi, W., Muennighoff, N., yu~Wang, H., Liu, H., Shi, Q., Siegel, Z.~S., Tang, M., Sun, R., Yoon, J., Arik, S.~O., Chen, D., and Yu, T.
\newblock {BRIGHT}: A realistic and challenging benchmark for reasoning-intensive retrieval.
\newblock \emph{arXiv preprint arXiv:2407.12883}, 2024.

\bibitem[Tafjord et~al.(2020)Tafjord, Mishra, and Clark]{tafjord2020proofwriter}
Tafjord, O., Mishra, B.~D., and Clark, P.
\newblock Proofwriter: Generating implications, proofs, and abductive statements over natural language.
\newblock \emph{arXiv preprint arXiv:2012.13048}, 2020.

\bibitem[Talmor \& Berant(2018)Talmor and Berant]{talmor2018web}
Talmor, A. and Berant, J.
\newblock The web as a knowledge-base for answering complex questions.
\newblock \emph{arXiv preprint arXiv:1803.06643}, 2018.

\bibitem[Tang \& Yang(2024)Tang and Yang]{tang2024multihop}
Tang, Y. and Yang, Y.
\newblock {MultiHop-RAG}: Benchmarking retrieval-augmented generation for multi-hop queries.
\newblock \emph{arXiv preprint arXiv:2401.15391}, 2024.

\bibitem[Trivedi et~al.(2022)Trivedi, Balasubramanian, Khot, and Sabharwal]{trivedi2022musique}
Trivedi, H., Balasubramanian, N., Khot, T., and Sabharwal, A.
\newblock {MuSiQue}: Multihop questions via single-hop question composition.
\newblock \emph{Transactions of the Association for Computational Linguistics}, 10:\penalty0 539--554, 2022.

\bibitem[Trivedi et~al.(2023)Trivedi, Balasubramanian, Khot, and Sabharwal]{trivedi2023interleaving}
Trivedi, H., Balasubramanian, N., Khot, T., and Sabharwal, A.
\newblock Interleaving retrieval with chain-of-thought reasoning for knowledge-intensive multi-step questions.
\newblock In \emph{Proceedings of the 61st Annual Meeting of the Association for Computational Linguistics (Volume 1: Long Papers)}, pp.\  10014--10037, 2023.

\bibitem[Vu et~al.(2023)Vu, Iyyer, Wang, Constant, Wei, Wei, Tar, Sung, Zhou, Le, et~al.]{vu2023freshllms}
Vu, T., Iyyer, M., Wang, X., Constant, N., Wei, J., Wei, J., Tar, C., Sung, Y.-H., Zhou, D., Le, Q., et~al.
\newblock {FreshLLMs}: Refreshing large language models with search engine augmentation.
\newblock \emph{arXiv preprint arXiv:2310.03214}, 2023.

\bibitem[Wang et~al.(2024)Wang, Chen, Fu, Liao, Zhang, Wu, Yu, Xu, Zhang, Luo, et~al.]{wang2024leave}
Wang, M., Chen, L., Fu, C., Liao, S., Zhang, X., Wu, B., Yu, H., Xu, N., Zhang, L., Luo, R., et~al.
\newblock Leave no document behind: Benchmarking long-context {LLMs} with extended multi-doc {QA}.
\newblock \emph{EMNLP}, 2024.

\bibitem[Wei et~al.(2022)Wei, Wang, Schuurmans, Bosma, Ichter, Xia, Chi, Le, and Zhou]{wei2022cot}
Wei, J., Wang, X., Schuurmans, D., Bosma, M., Ichter, B., Xia, F., Chi, E., Le, Q., and Zhou, D.
\newblock Chain-of-thought prompting elicits reasoning in large language models.
\newblock \emph{Advances in neural information processing systems}, 35:\penalty0 24824--24837, 2022.

\bibitem[Welbl et~al.(2018)Welbl, Stenetorp, and Riedel]{welbl2018constructing}
Welbl, J., Stenetorp, P., and Riedel, S.
\newblock Constructing datasets for multi-hop reading comprehension across documents.
\newblock \emph{Transactions of the Association for Computational Linguistics}, 6:\penalty0 287--302, 2018.

\bibitem[Weston et~al.(2015)Weston, Bordes, Chopra, Rush, Van~Merri{\"e}nboer, Joulin, and Mikolov]{weston2015towards}
Weston, J., Bordes, A., Chopra, S., Rush, A.~M., Van~Merri{\"e}nboer, B., Joulin, A., and Mikolov, T.
\newblock Towards {AI-complete} question answering: A set of prerequisite toy tasks.
\newblock \emph{arXiv preprint arXiv:1502.05698}, 2015.

\bibitem[Yang et~al.(2018)Yang, Qi, Zhang, Bengio, Cohen, Salakhutdinov, and Manning]{yang2018hotpotqa}
Yang, Z., Qi, P., Zhang, S., Bengio, Y., Cohen, W.~W., Salakhutdinov, R., and Manning, C.~D.
\newblock {HotpotQA}: A dataset for diverse, explainable multi-hop question answering.
\newblock \emph{arXiv preprint arXiv:1809.09600}, 2018.

\bibitem[Yao et~al.(2022)Yao, Zhao, Yu, Du, Shafran, Narasimhan, and Cao]{yao2022react}
Yao, S., Zhao, J., Yu, D., Du, N., Shafran, I., Narasimhan, K., and Cao, Y.
\newblock {ReAct}: Synergizing reasoning and acting in language models.
\newblock \emph{arXiv preprint arXiv:2210.03629}, 2022.

\bibitem[Yao et~al.(2024)Yao, Shinn, Razavi, and Narasimhan]{yao2024tau}
Yao, S., Shinn, N., Razavi, P., and Narasimhan, K.
\newblock {$\tau$}-bench: A benchmark for tool-agent-user interaction in real-world domains.
\newblock \emph{arXiv preprint arXiv:2406.12045}, 2024.

\bibitem[Ye et~al.(2022)Ye, Zhang, Chen, and Chen]{ye2022generative}
Ye, H., Zhang, N., Chen, H., and Chen, H.
\newblock Generative knowledge graph construction: A review.
\newblock \emph{arXiv preprint arXiv:2210.12714}, 2022.

\bibitem[Zellers et~al.(2019)Zellers, Holtzman, Bisk, Farhadi, and Choi]{zellers2019hellaswag}
Zellers, R., Holtzman, A., Bisk, Y., Farhadi, A., and Choi, Y.
\newblock {HellaSwag}: Can a machine really finish your sentence?
\newblock \emph{arXiv preprint arXiv:1905.07830}, 2019.

\bibitem[Zhang et~al.(2024)Zhang, Chen, Hu, Xu, Chen, Hao, Han, Thai, Wang, Liu, et~al.]{zhang2024infty}
Zhang, X., Chen, Y., Hu, S., Xu, Z., Chen, J., Hao, M.~K., Han, X., Thai, Z.~L., Wang, S., Liu, Z., et~al.
\newblock $\infty${Bench}: Extending long context evaluation beyond {100K} tokens.
\newblock \emph{ACL}, 2024.

\bibitem[Zhou et~al.(2023)Zhou, Xu, Zhu, Zhou, Lo, Sridhar, Cheng, Ou, Bisk, Fried, Alon, and Neubig]{zhou2023webarena}
Zhou, S., Xu, F.~F., Zhu, H., Zhou, X., Lo, R., Sridhar, A., Cheng, X., Ou, T., Bisk, Y., Fried, D., Alon, U., and Neubig, G.
\newblock {WebArena}: A realistic web environment for building autonomous agents.
\newblock \emph{arXiv preprint arXiv:2307.13854}, 2023.

\bibitem[Zhu et~al.(2023)Zhu, Xue, Chen, Zhou, Tang, Schuurmans, and Dai]{zhu2023large}
Zhu, Z., Xue, Y., Chen, X., Zhou, D., Tang, J., Schuurmans, D., and Dai, H.
\newblock Large language models can learn rules.
\newblock \emph{arXiv preprint arXiv:2310.07064}, 2023.

\bibitem[Zhuang et~al.(2023)Zhuang, Yu, Wang, Sun, and Zhang]{zhuang2023toolqa}
Zhuang, Y., Yu, Y., Wang, K., Sun, H., and Zhang, C.
\newblock {ToolQA}: A dataset for {LLM} question answering with external tools.
\newblock \emph{Advances in Neural Information Processing Systems}, 36:\penalty0 50117--50143, 2023.

\end{thebibliography}
